\documentclass{article}

\usepackage{microtype}
\usepackage{graphicx}
\usepackage{subfigure}
\usepackage{booktabs} 

\usepackage{hyperref}

\usepackage[accepted]{icml2023}

\usepackage{amsmath}
\usepackage{amssymb}
\usepackage{mathtools}
\usepackage{amsthm}
\usepackage{katsuki}
\usepackage{bm}
\usepackage{nicefrac}

\usepackage{algorithm}
\usepackage{algorithmic}
\renewcommand{\algorithmicrequire}{\textbf{input:}}
\renewcommand{\algorithmicensure}{\textbf{output:}}

\usepackage[capitalize,noabbrev]{cleveref}

\theoremstyle{plain}
\newtheorem{theorem}{Theorem}[section]
\newtheorem{proposition}[theorem]{Proposition}
\newtheorem{lemma}[theorem]{Lemma}

\theoremstyle{definition}

\newtheorem{assumption}[theorem]{Assumption}
\newtheorem{condition}[theorem]{Condition}
\theoremstyle{remark}

\usepackage[textsize=tiny]{todonotes}

\icmltitlerunning{Regression with Sensor Data Containing Incomplete Observations}

\begin{document}

\twocolumn[
\icmltitle{Regression with Sensor Data Containing Incomplete Observations}




\begin{icmlauthorlist}
\icmlauthor{Takayuki Katsuki}{comp}
\icmlauthor{Takayuki Osogami}{comp}
\end{icmlauthorlist}

\icmlaffiliation{comp}{IBM Research - Tokyo, Tokyo, Japan}

\icmlcorrespondingauthor{Takayuki Katsuki}{kats@jp.ibm.com}

\icmlkeywords{regression, unlabeled data, health informatics, sensor data analytics, non-invasive sensing, weakly supervised learning, weakly supervised regression}

\vskip 0.3in
]




\begin{abstract}
This paper addresses a regression problem in which output label values are the results of sensing the magnitude of a phenomenon. A low value of such labels can mean either that the actual magnitude of the phenomenon was low or that the sensor made an incomplete observation. This leads to a bias toward lower values in labels and the resultant learning because labels may have lower values due to incomplete observations, even if the actual magnitude of the phenomenon was high. Moreover, because an incomplete observation does not provide any tags indicating incompleteness, we cannot eliminate or impute them. To address this issue, we propose a learning algorithm that explicitly models incomplete observations corrupted with an asymmetric noise that always has a negative value. We show that our algorithm is unbiased as if it were learned from uncorrupted data that does not involve incomplete observations. We demonstrate the advantages of our algorithm through numerical experiments.
\end{abstract}


\section{Introduction}


This paper addresses a regression problem for predicting the magnitude of a phenomenon when an observed magnitude involves a particular measurement error. The magnitude typically represents \emph{how large a phenomenon is or how strong the nature of the phenomenon is}. Such examples of predicting the magnitude are found in several application areas, including pressure, vibration, and temperature~\citep{vandal2017deepsd,shi2017deep,wilby2004guidelines,tanaka2019refining}. In medicine and healthcare, the magnitude may represent pulsation, respiration, or body movements~\citep{inan2009non,nukaya2010noninvasive,lee2016physiological,alaziz2016motion,alaziz2017motiontree,carlson2018pilot}.

More specifically, we learn a regression function to predict the \emph{label} representing the magnitude of a phenomenon from \emph{explanatory variables}, where the label is observed with a sensor and is not necessarily in agreement with the actual magnitude.
We use the term ``label'' even though we address the regression problem; a label is thus real-valued in this paper.

For example, the body movements of a patient are typically measured with an intrusive sensor attached to the chest or wrist. If we can learn a reliable regression function that predicts the magnitude of the body movements (label) from the values of non-intrusive bed sensors (explanatory variables)~\citep{mullaney1980wrist,webster1982activity,cole1992automatic,tryon2013activity}, we can replace intrusive sensors with non-intrusive ones, which will reduce the burden on patients.

\begin{figure}
    \centering
    \includegraphics[width=80mm]{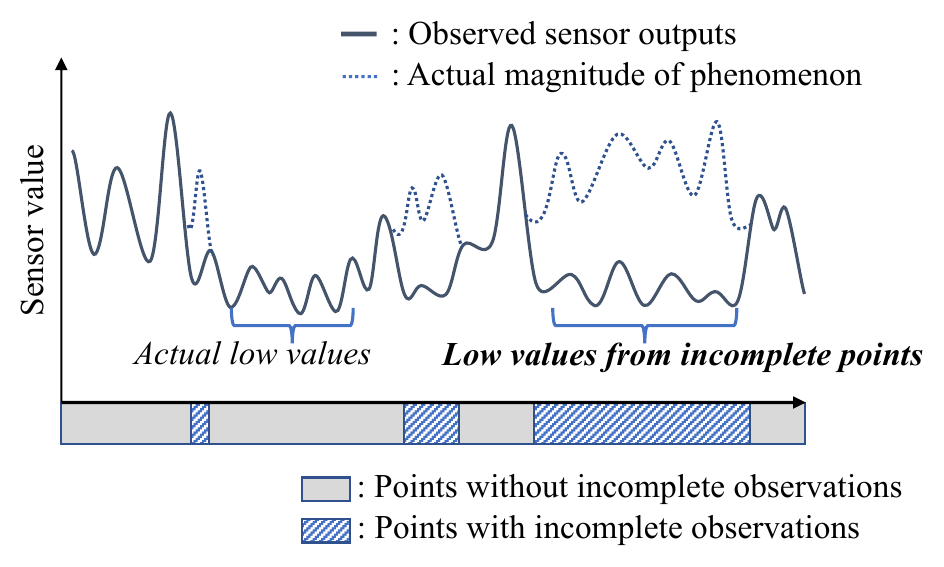}
    \caption{Sensor values with incomplete observations. Low values can mean either actual low magnitude or incomplete observation.}
    \label{FigProblem}
\end{figure}

Although the sensors that measure the label generally have high accuracy, they often make incomplete observations, and \emph{such incomplete observations are recorded as low values instead of missing values}. This leads to the particular challenge illustrated in Fig.~\ref{FigProblem}, where a low value of the label can mean either that the actual magnitude of the phenomenon was low or that the sensor made an incomplete observation, and there are no clues to distinguish between the two.

Such incomplete observations are prevalent when measuring the magnitude of a phenomenon. For example, the phenomenon may be outside the coverage of a sensor, or the sensing system may experience temporary mechanical failures. In the case of body movements, the sensor may be temporarily detached from the chest or wrist. In all cases, the sensor keeps recording low values, even though the actual magnitude may be high, and no tag indicating incompleteness is provided.

This incomplete observation is particularly severe in the sensor for the label, not for the explanatory variables. The sensor for the label is often single-source and has narrow data coverage because it is intrusive or because it is expensive to produce highly accurate observations.
For example, chest or wrist sensors focus on the movements of a local body part with high accuracy and often miss movements outside their coverage, such as those of parts located far from where the sensor is attached.
At most, a single intrusive sensor can be attached to avoid burdening the patient.
In contrast, the sensors for explanatory variables are usually multi-source and provide broader data coverage: for example, multiple sensors can be attached on various places of a bed and globally monitor the movements of all body parts, albeit with low accuracy.

One cannot simply ignore the problem that the observations of labels may be incomplete because estimated regression functions naively trained on such data would be severely biased toward lower values regardless of the amount of training data. This bias comes from the fact that incomplete observations always have lower values than the actual magnitude, and they frequently occur on label sensors, whereas explanatory variables are observed completely.

Unfortunately, since we cannot identify which observations are incomplete, we cannot eliminate or impute them by using existing methods that require the identification of incomplete observations. Such methods include thresholding, missing value detection~\citep{pearson2006problem,qahtan2018fahes}, imputation~\citep{enders2010applied,smieja2018processing,ma2019missing,sportisse2020estimation}, and semi-supervised regression~\citep{zhou2005semi,zhu2009introduction,jean2018semi,zhou2019graph}.

The problems of incomplete observations also cannot be solved with robust regression~\citep{huber1964robust,narula1982minimum,draper1998applied,wilcox2012introduction}, which takes into account the possibility that the observed labels contain outliers.
While robust regression is an established approach and state-of-the-art for corrupted labels in regression, it assumes that the noise is unbiased. Since incomplete observations induce noise that is severely biased toward lower values, robust regression methods still produce regression functions that are biased toward lower values when the fraction of incomplete observations exceeds its tolerance.

In this paper, to mitigate the bias toward lower values, we explicitly assume the existence of asymmetric noise in labels from incomplete observations, which always has a negative value, in addition to the ordinary symmetric noise, i.e., \emph{asymmetric label corruption}. We refer to data with the asymmetric label corruption as \emph{asymmetrically corrupted data}. We then formulate a regression problem from our asymmetrically corrupted data and design a principled learning algorithm.

By explicitly modeling the asymmetric label corruption, we derive a learning algorithm that has a rather bold feature: it ignores the labels that have relatively low values (lower-side labeled data). In other words, our algorithm utilizes the data whose labels have relatively high values (upper-side labeled data) and the data whose labels are ignored (unlabeled data). Hence, we refer to our algorithm as \emph{upper and unlabeled} regression (U2 regression). This aligns with the intuition that the labels with low values are particularly unreliable, as those low values may be due to incomplete observations.

Our main result is that U2 regression
produces a regression function that is, under some technical assumptions, unbiased and consistent with the one that could be produced from uncorrupted data (i.e., without incomplete observations).
This counterintuitive result is achieved by considering a specific class of loss functions and deriving their gradient as a form requiring only upper-side labeled data and unlabeled data. The gradient can be computed with only the reliable parts in asymmetrically corrupted data.
We prove that this gradient is asymptotically equivalent to the gradient that is computed with the uncorrupted data.
The main novelty of our approach is thus in the loss function, and we will empirically demonstrate the effectiveness of the proposed class of loss functions over common existing loss functions in dealing with asymmetrically corrupted data in synthetic and six real-world regression tasks.

\paragraph{Contributions.}
\begin{itemize}
  \item We formulate a novel problem of learning a regression function for a sensor magnitude with asymmetrically corrupted data. This is vital for applications where the sensor is susceptible to unidentifiable incomplete observations.
  \item We derive an unbiased and consistent learning algorithm (U2 regression) for this problem with the new class of loss functions.
  \item Extensive experiments on synthetic and six real-world regression tasks including a real use case for healthcare demonstrate the effectiveness of the proposed method.
\end{itemize}

\section{Regression from Asymmetrically Corrupted Data}
Our goal is to derive a learning algorithm with asymmetrically corrupted data (due to incomplete observations) in a manner that is unbiased and consistent with the one that uses uncorrupted data (without involving incomplete observations).
In Section~\ref{sec:approach:without}, we examine the regression problem that uses the uncorrupted data, and in Section~\ref{sec:approach:with}, we formulate learning from the asymmetrically corrupted data.


\subsection{Regression Problem from Uncorrupted Data}
\label{sec:approach:without}
Let $\bx \in \mathbb{R}^D (D \in \mathbb{N})$ be a $D$-dimensional explanatory variable and $y \in \mathbb{R}$ be a real-valued label.
We assume that, without incomplete observations, $y$ is observed in accordance with
\begin{align}
    \label{observation_process_unbiased}
    y&=f^*(\bx) + \epsilon_{\mathrm{s}},
\end{align}
where $f^*$ is the oracle regressor and $\epsilon_{\mathrm{s}}$ is the symmetric noise with $0$ as the center, such as additive white Gaussian noise (AWGN).

We learn a regression function $f(\bx)$ that computes the value of the estimation of a label $\hy$ for a newly observed $\bx$ as $\hy = f(\bx)$.
The optimal regression function $\hf$ is given by
\begin{align}
    \label{objective}
    \hf \equiv \argmin_{f\in \calF} \mathcal{L}(f),
\end{align}
where $\calF$ is a hypothesis space for $f$, and $\mathcal{L}(f)$ is the expected loss when the regression function $f(\bx)$ is applied to data $(\bx, y)$, distributed in accordance with an underlying distribution $p(\bx,y)$:
\begin{align}
    \label{Loss}
    \mathcal{L}(f) &\equiv \mathbb{E}_{p(\bx,y)}[L(f(\bx),y)],
\end{align}
where $\mathbb{E}_{p}[\bullet]$ denotes the expectation over the distribution $p$, and $L(f(\bx),y)$ is the loss function between $f(\bx)$ and $y$, e.g., the squared loss, $L(f(\bx), y)=\|f(\bx)-y\|^2$.
The expectation $\mathbb{E}_{p(\bx,y)}$ can be estimated by computing a sample average for the training data $\bm{\calD}\equiv\{ (\bx_n, y_n) \}^N_{n=1}$, which is $N$ pairs of explanatory variables and labels.

\subsection{Regression Problem from Asymmetrically Corrupted Data}
\label{sec:approach:with}
In this paper, we consider a scenario in which we only have access to the asymmetrically corrupted data $\bm{\calD}'\equiv\{ (\bx_n, y'_n) \}^N_{n=1}$, where a label $y'$ may be corrupted due to incomplete observations.
A corrupted label $y'$ is observed from the uncorrupted $y$ with an asymmetric negative-valued noise, $\epsilon_{\mathrm{a}}$:
\begin{align}
    \label{observation_process}
    y' &= y + \epsilon_{\mathrm{a}},
\end{align}
where $\epsilon_{\mathrm{a}}$ always has a random negative value, which implies $y'\leq y$.

We seek to learn a regression function $f(\bx)$ \emph{in an unbiased and consistent manner} (i.e., to find the solution for Eq.~\eqref{objective}) by using only $\bm{\calD}'$. Although AWGN can be handled even when we use a naive regression method such as least squares, the asymmetric noise $\epsilon_{\mathrm{a}}$, which always has a negative value, is problematic.

\begin{figure}
    \centering
    \includegraphics[width=80mm]{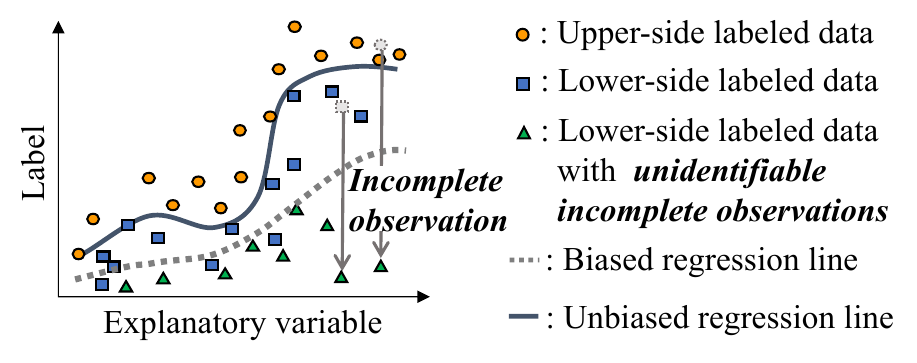}
    \caption{Asymmetrically corrupted data. Labels for incomplete observations, depicted as triangles, become lower than those of typical observations, depicted as circles or squares.}
    \label{FigData}
\end{figure}

Intuitively, the asymmetric noise $\epsilon_{\mathrm{a}}$ makes \emph{lower-side labeled data} particularly unreliable and inappropriate for learning while keeping \emph{upper-side labeled data} reliable, where the upper-side labeled data refers to the data $\{(\bx,y)\}$ whose label is above the regression line (i.e., $f(\bx)\leq y$) and the lower-side labeled data refers to the data whose label is below the regression line (i.e., $y < f(\bx)$). The regression line represents the estimation of a regression function. Figure~\ref{FigData} illustrates this as a scatter plot of the value of the label against the value of an explanatory variable. Here, the data with incomplete observations appear only in the lower side of the regression line because $\epsilon_{\mathrm{a}}$ makes observations have lower label values than those of typical observations, where the regression line represents such typical observations. This asymmetry leads to biased learning compared to learning from the uncorrupted data without incomplete observations.

To address the asymmetric noise $\epsilon_{\mathrm{a}}$ and its resultant bias, we formalize the assumption on the observation process for the asymmetrically corrupted data $\bm{\calD}'$ and derive a lemma representing the nature of $\bm{\calD}'$. Then, we propose a learning algorithm based on the lemma in the next section.

We assume that $\bm{\calD}'$ has enough information to estimate $f$ and that the asymmetric noise $\epsilon_{\mathrm{a}}$ is significant enough compared to the symmetric noise $\epsilon_{\mathrm{s}}$, which are necessary assumptions to make the learning problem solvable. Formally, the observation processes of $\bm{\calD}$ and $\bm{\calD}'$ are characterized as follows.
\newcommand{\indep}{\mathop{\perp\!\!\!\!\perp}}
\begin{assumption}
    \label{assumption_obsprocess}
    Assume $\epsilon_{\mathrm{s}} \indep f^*(\bx)$, $\mathbb{E}_{p(\epsilon_{\mathrm{s}})}[\epsilon_{\mathrm{s}}] = 0$; $\epsilon_{\mathrm{a}} \indep f^*(\bx)$, $\epsilon_{\mathrm{a}}\leq 0$ almost surely (a.s.); $2 |\epsilon_\mathrm{s}| < |\epsilon_\mathrm{a}|$ a.s. when $\epsilon_{\mathrm{a}}<0$; and $\{ (\bx_n, y_n, y'_n) \}^N_{n=1}$ are i.i.d. observations in accordance with Eqs.~\eqref{observation_process_unbiased} and~\eqref{observation_process}.
\end{assumption}

We then have the following lemma, which shows that $\epsilon_{\mathrm{a}}$ does not change the expectation for our upper-side labeled data ($f(\bx)\leq y'$) before and after adding $\epsilon_{\mathrm{a}}$.
\begin{lemma}
    \label{lemma_obsprocess_E}
    Let $\calF' \equiv \{f\in\calF : |f(\bx)-f^*(\bx)| \leq |\epsilon_{\mathrm{s}}| \mbox{ a.s.}\}$. When $f\in\calF'$, the following holds under Assumption~\ref{assumption_obsprocess}:
    \begin{align}
        \label{consistent_E_upper2}
        \mathbb{E}_{p(\bx,y'\|f(\!\bx\!)\leq y')}[G(\bx,y')] &= \mathbb{E}_{p(\bx,y|f(\!\bx\!)\leq y)}[G(\bx,y)]
    \end{align}
    for any function $G:\mathbb{R}^D\times\mathbb{R}\to\mathbb{R}$ as long as the expectations exist.
\end{lemma}
\begin{proof}
We outline a proof here and provide a complete one in Appendix~\ref{ap_pr_lemma_obsprocess_E}.
We first show that $\epsilon_{\mathrm{a}}$ does not change the distribution for upper-side labeled data ($f^*(\bx)\leq y'$) on the basis of the oracle regression function $f^*$ before and after adding $\epsilon_{\mathrm{a}}$, i.e., $\epsilon_{\mathrm{a}}=0$ when $f^*(\bx)\leq y'$. With the condition $f\in\calF'$, we can further prove that $\epsilon_{\mathrm{a}}=0$ when $f(\bx)\leq y'$, which is for upper-side labeled data on the basis of $f$. This establishes $p(x,y'|f(\bx)\leq y') = p(x,y|f(\bx)\leq y)$ and implies Lemma~\ref{lemma_obsprocess_E}.
\end{proof}

\noindent The condition parts of the conditional distributions in Eq.~\eqref{consistent_E_upper2} represent the relationships between labels and the estimations of the regression function $f$, e.g., $p(\bx,y|f(\bx)\leq y)$ is the distribution of $x$ and $y$ when $y$ is higher than what is given by $f$. The condition $f\in\calF'$ represents our natural expectation that $f$ well approximates $f^*$.

Lemma~\ref{lemma_obsprocess_E} shows that our upper-side labeled data ($f(\bx)\leq y'$) is still reliable for regression. In the next section, we derive an unbiased learning algorithm based on this lemma.

\section{U2 Regression}
We seek to find the minimizer of the objective $\mathcal{L}(f)$ in Eq.~\eqref{objective} {\it from the asymmetrically corrupted data $\bm{\calD}'$}. To this end, we propose a gradient that relies only on the knowledge of the distribution of the corrupted data $p(\bx,y')$ but is still equivalent to the gradient of $\mathcal{L}(f)$, which depends on the distribution of the uncorrupted data $p(\bx,y)$.
Specifically, based on Lemma~\ref{lemma_obsprocess_E}, we will rewrite the gradient based on $p(\bx,y)$ into one that only requires $p(\bx,y')$.

\subsection{Gradient for Regression from Asymmetrically Corrupted Data}
Here, we minimize $\mathcal{L}(f)$ with the gradient descent.
At step $t+1$ in the gradient descent, the gradient of $\mathcal{L}(f)$ with respect to the parameters $\btheta$ of $f$ is represented with a regression function, $f_t$, which is estimated at step $t$, as follows:
\begin{align}
    \label{gradient_new}
    &\nabla \mathcal{L}(f_t) \equiv \mathbb{E}_{p(\bx,y)}[\nabla L(f_t(\bx),y)],\\\nonumber
    &\mathrm{where}~~~~
    \nabla L(f_t(\bx),y) \equiv\Big.\frac{\partial L(f(\bx),y)}{\partial \btheta}\Big|_{f=f_t}.
\end{align}
Note that this holds for any step in the gradient descent. When $t=0$, $f_0$ is the initial value of $f$, and when $t=\infty$, we suppose $f_{\infty}=\hf$.
We can decompose $\nabla \mathcal{L}(f_t)$ as
\begin{align}
    \label{decomposedGradient_new}
    &\nabla \mathcal{L}(f_t) =p(f_t(\bx)\leq y) \mathbb{E}_{p(\bx,y|f_t(\bx)\leq y)}[\nabla L(f_t(\bx),y)]\nonumber\\
    &~~~~~~~~+ p(y<f_t(\bx)) \mathbb{E}_{p(\bx,y|y<f_t(\bx))}[\nabla L(f_t(\bx),y)].
\end{align}

We will introduce and use the class of loss functions whose gradient can depend on $f(\bx)$ but not on $y$ when $y<f(\bx)$; thus, when $y<f(\bx)$, we write $\nabla L(f(\bx),y)$ as $\bg(f(\bx))$ to emphasize this independence. Formally,
\begin{condition}
    \label{condition_L}
   For $y<f(\bx)$, let $\bg(f(\bx))$ be defined as $\nabla L(f(\bx),y)$. Then $\bg(f(\bx))$ depends only on $f(\bx)$ and is conditionally independent of $y$ given $f(\bx)$.
\end{condition}
\noindent Such common losses that satisfy Condition~\ref{condition_L} include the absolute loss and pinball loss, which are respectively used in least absolute regression and quantile regression~\citep{lee2016physiological,yeung2002reverse,wang2005robust,srinivas2020prediction}.
For example, the gradient of the absolute loss is
\begin{align}
    \label{gradient_abs}
    \frac{\partial |f(\bx)-y|} {\partial \btheta} = \frac{\partial f(\bx)} {\partial \btheta}\qquad\mathrm{when}~y<f(\bx),
\end{align}
which does not depend on the value of $y$ but only on $f(\bx)$.

The class of loss functions satisfying Condition~\ref{condition_L} is broad, since Condition~\ref{condition_L} only specifies the loss function for lower-side labeled data ($y<f(x)$). The upper-side loss can be an arbitrary function. In our experiments, we will actually use squared loss for upper-side labeled data ($f(x) \leq y$), which violates Condition~\ref{condition_L}, and absolute loss for lower-side labeled data ($y<f(x)$).

We now propose a gradient that does not rely on the knowledge of $p(\bx,y)$ but instead uses only $p(\bx,y')$. Namely,
\begin{align}
    \label{finalGradient_new}
    \nabla \mathcal{\tilde{L}}(f_t)& \equiv p(f_t(\bx)\leq y) \mathbb{E}_{p(\bx,y'|f_t(\bx)\leq y')}\Big[\nabla L(f_t(\bx),y)\Big]\nonumber\\
     &+ \mathbb{E}_{p(\bx)}\Big[\bg(f_t(\bx))\Big]\nonumber\\
     &- p(f_t(\bx)\leq y) \mathbb{E}_{p(\bx|f_t(\bx)\leq y')}\Big[\bg(f_t(\bx))\Big].
\end{align}
In Section~\ref{sec:unbiasedness}, we will formally establish the equivalence between the gradient in Eq.~\eqref{finalGradient_new} and that in Eq.~\eqref{gradient_new} under our assumptions. Note that, in the second and third terms of Eq.~\eqref{finalGradient_new}, we apply expectations over $p(\bx)$ and $p(\bx|f_t(\bx)\leq y')$ to $\bg(f(\bx))$, even though $\bg(f(\bx))$ is defined only for $y<f(\bx)$. This is tractable due to the nature of $\bg(f(\bx))$, which does not depend on the value of $y$.

Since the expectations in Eq.~\eqref{finalGradient_new} only depend on $\bx$ and $y'$, they can be estimated by computing a sample average for our asymmetrically corrupted data $\bm{\calD}'$ as
\begin{align}
    \label{empiricalgradient}
    &\nabla \mathcal{\hL}(f_t)=\frac{\pi_{\mathrm{up}}}{n_{\mathrm{up}}} \bigg[\sum_{(\bx,y) \in \{\bX_{\mathrm{up}},\by'_{\mathrm{up}}\}} \nabla L(f_t(\bx),y)\bigg]\\\nonumber
    &~~~~~~~~~~+ \frac{1}{N}\bigg[\sum_{\bx \in \bX_{\mathrm{un}}} \bg(f_t(\bx)) \bigg] - \frac{\pi_{\mathrm{up}}}{n_{\mathrm{up}}} \bigg[\sum_{\bx \in \bX_{\mathrm{up}}} \bg(f_t(\bx))\bigg],
\end{align}
where $\{\bX_{\mathrm{up}},\by'_{\mathrm{up}}\}$ represents the set of coupled pairs of $\bx$ and $y'$ in the upper-side labeled sample set, $\{x,y'\colon f_t(\bx)\leq y' \}$, in $\bm{\calD}'$; $\bX_{\mathrm{un}}$ is a sample set of $\bx$ in $\bm{\calD}'$ ignoring labels $y'$; $n_{\mathrm{up}}$ is the number of samples in the upper-side labeled set; and $\pi_{\mathrm{up}}$ is $\pi_{\mathrm{up}}\equiv p(f_t(\bx)\leq y)$. Note that $\pi_{\mathrm{up}}$ depends on the current estimation of the function $f_t$ and the label $y$ with complete observation. Thus, it changes at each step of the gradient descent, and we cannot determine its value in a general way. In this paper, we propose treating $\pi_{\mathrm{up}}$ as a hyperparameter and optimize it with the grid search based on the validation set. In our experiments, we will demonstrate that this simple approach can work effectively and robustly in practice.

As we will show in Section~\ref{sec:unbiasedness}, we can use Eq.~\eqref{empiricalgradient} to design an algorithm that gives an unbiased and consistent regression function.
By using the gradient in Eq.~\eqref{empiricalgradient}, we can optimize Eq.~\eqref{objective} and learn the regression function with only upper-side labeled samples and unlabeled samples from $\bm{\calD}'$ independent of lower-side labels. This addresses the problem that our lower-side labeled data is particularly unreliable and leads to overcoming the bias that stems from this unreliable part of the data. We refer to our algorithm as \emph{upper and unlabeled} regression~(U2 regression).

See Appendix~\ref{ap_implementation} for the specific implementation of the algorithm based on stochastic optimization.
The gradient in Eq.~\eqref{empiricalgradient} can be interpreted in an intuitive manner. The first term has the effect of minimizing the upper-side loss. Recall that the upper-side data are not affected by the asymmetric noise under our assumptions from Lemma~\ref{lemma_obsprocess_E}. Thus, U2 regression seeks to learn the regression function $f$ on the basis of this reliable upper-side data. Note that the first term becomes zero when all of the data points are below $f$ (i.e., $y' \leq f_t(\bx), \forall (x,y')\in\calD'$), since $\{\bX_{\mathrm{up}}, \by'_{\mathrm{up}}\big\}$ then becomes empty. The second term thus has the effect of pushing down $f$ at all of the data points so that some data points are above $f$. Meanwhile, the third term partially cancels out this effect of the second term for the upper-side data to control the balance between the first and second terms.

\subsection{Unbiasedness and Consistency of Gradient}
\label{sec:unbiasedness}
U2 regression is the learning algorithm based on the gradient $\nabla \mathcal{\hL}(f_t)$ in Eq.~\eqref{empiricalgradient} and uses only asymmetrically corrupted data $\bm{\calD}'$. The use of $\nabla \mathcal{\hL}(f_t)$ can be justified as follows:
\begin{proposition}
    \label{proposition1}
    Suppose that Assumption~\ref{assumption_obsprocess} holds and the loss function $L(f(\bx),y)$ satisfies Condition~\ref{condition_L}. Then, the gradient $\nabla \mathcal{\tilde{L}}(f_t)$ in Eq.~\eqref{finalGradient_new} and its empirical approximation $\nabla \mathcal{\hL}(f_t)$ in Eq.~\eqref{empiricalgradient} are unbiased and consistent with the gradient $\nabla \mathcal{L}(f_t)$ in Eq.~\eqref{gradient_new} a.s.
\end{proposition}
\begin{proof}
We outline a proof here and provide a complete one in Appendix~\ref{ap_pr_proposition1}.
First, we rewrite the decomposed gradient $\nabla \mathcal{L}(f_t)$ in Eq.~\eqref{decomposedGradient_new} into a gradient that only contains the expectations over $p(\bx,y|f_t(\bx)\leq y)$ and $p(\bx)$ with Condition~\ref{condition_L}.
Then, we apply Lemma~\ref{lemma_obsprocess_E} to the gradient, and it becomes identical to Eq.~\eqref{finalGradient_new}.
\end{proof}

\noindent In other words, U2 regression asymptotically produces the same result as the learning algorithm based on the gradient $\nabla \mathcal{L}(f_t)$ in Eq.~\eqref{gradient_new}, which requires the uncorrupted data without incomplete observations, $\bm{\calD}$. The convergence rate of U2 regression is of the order $\mathcal{O}_p(\nicefrac{1}{\sqrt{n_{\mathrm{up}}}}+ \nicefrac{1}{\sqrt{N}})$ in accordance with the central limit theorem~\citep{chung1968course}, where $\mathcal{O}_p$ denotes the order in probability.

We further justify our approach of having the specific form of Eq.~\eqref{finalGradient_new} by showing how a straightforward variant that uses $\bm{\calD}'$ as if it does not involve incomplete observations (i.e., $p(\bx,y) \approx p(\bx,y')$) can fail for our problem.
To this end, we introduce an additional assumption on the observation process:
\begin{assumption}
    \label{additional_es}
    Assume $\epsilon_{\mathrm{a}} \indep \bx$.
\end{assumption}

Then, we have
\begin{lemma}
    \label{naive}
    Let $\nabla \check{\mathcal{L}}(f_t)$ be a variant of the gradient in Eq.~\eqref{decomposedGradient_new} replacing $p(\bx,y)$ with $p(\bx,y')$, $\delta$ be the difference between the expectations of the gradients in the upper side and the lower side $\delta \equiv \big|\mathbb{E}_{p(\bx,y|f(\bx)\leq y)}[\nabla L(f(\bx),y)]$ $- \mathbb{E}_{p(\bx,y|y<f(\bx))}[\nabla L(f(\bx),y)] \big|$, $\eta\in[0, 1]$ be the probability of being $0\leq \epsilon_{\mathrm{s}}$, and $\xi\in[0, 1]$ be the probability of $\epsilon_{\mathrm{a}}=0$.
    Then, $\nabla \check{\mathcal{L}}(f_t)$ is not consistent with the gradient $\nabla \mathcal{L}(f_t)$ in Eq.~\eqref{gradient_new} a.s., and the difference (bias) between them at step $t+1$ in the gradient descent is
     \begin{align}
         \frac{\eta(1-\eta)(1-\xi)}{1-\eta\xi}\delta \leq |\nabla \check{\mathcal{L}}(f_t)-\nabla \mathcal{L}(f_t)|.
     \end{align}
\end{lemma}
\begin{proof}
We outline a proof here and provide a complete one in Appendix~\ref{ap_pr_naive}.
We first show that the bias $|\nabla \check{\mathcal{L}}(f_t)-\nabla \mathcal{L}(f_t)|$ can be represented by the difference between the expectation of $\bg(f_t(\bx))$ with the upper-side data and that with the lower-side data, which can be written by $\delta$. The bias also has a coefficient that contains the proportions for the lower-side data and the original upper-side data mixed into the lower side due to incomplete observations. These values can be written by $\eta$ and $\xi$ from their definitions.
\end{proof}

Lemma~\ref{naive} shows that the bias caused by the asymmetric label corruption becomes severe when there is a large difference between the expectations of the gradients in the upper side and the lower side. $\delta$ is usually higher than zero because $\delta = 0$ implies that there is no difference between the expectations of the gradients in the upper and lower sides or that both of the expectations are zero. Furthermore, a larger $1-\xi=p(\epsilon_{\mathrm{a}}<0)$ makes the bias more significant, which agrees with the intuition that as the proportion of incomplete observations increases, the problem becomes more difficult.

\section{Experiments}
We now evaluate the proposed method through numerical experiments. We first introduce the baselines to be compared and then present the experimental results to demonstrate the effectiveness of our unbiased learning.
\begin{figure*}[t]
	\centering
	\includegraphics[width=170mm]{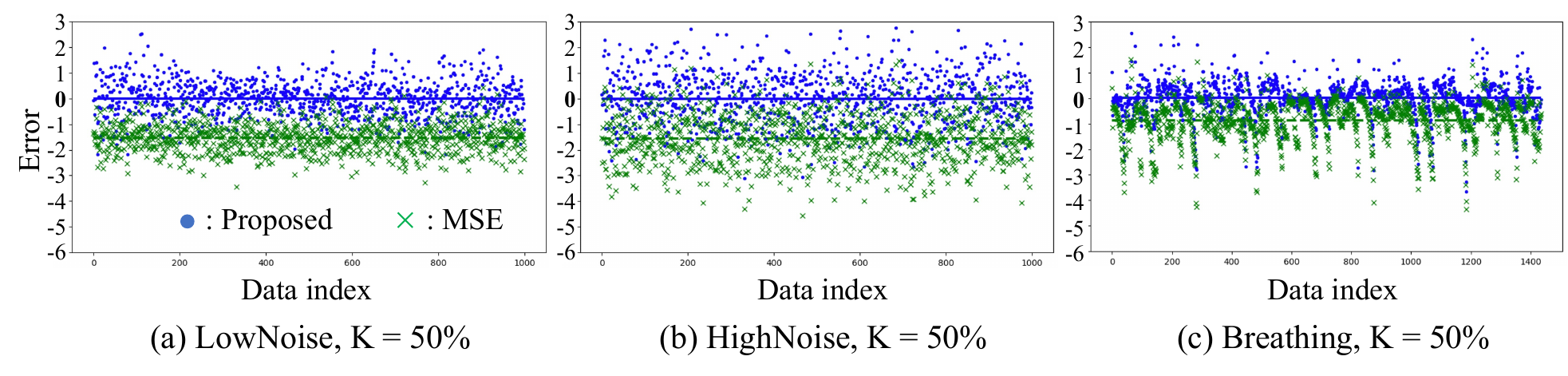}
	\caption{Errors (predicted value minus true value) by proposed method (blue) and by \texttt{MSE} (green) for three tasks: (a) \textbf{LowNoise}, (b) \textbf{HighNoise}, and (c) \textbf{Breathing}, with $K=50 \%$ of incomplete training samples. Error of each data point is shown by dots (for proposed method) or crosses (for \texttt{MSE}), and average error is shown by solid line (proposed method) or dashed line (\texttt{MSE}).}
	\label{FigResultsSD}
\end{figure*}

\subsection{Baselines}
Recall that the novelty of the proposed approach lies in the unbiased gradient in Eq.~\eqref{empiricalgradient}, which is derived from the new class of loss functions in Eq.~\eqref{finalGradient_new} with Condition~\ref{condition_L}. The objective of our experiments is thus to validate the effectiveness of this new class of loss functions and the corresponding gradients against common loss functions in the literature.
Specifically, we compare the proposed method with mean squared error (MSE), mean absolute error (MAE), and Huber losses~\citep{huber1964robust,narula1982minimum,wilcox2012introduction}. In terms of robust loss functions in regression, MAE and Huber losses are considered the de facto standard and state-of-the-art in many studies and libraries.
We use the same model and optimization method with all of the loss functions under consideration, and hence the only difference between the proposed method and the baselines is the loss functions. Since the loss function uniquely determines the baseline, we refer to each baseline method as \texttt{MSE}, \texttt{MAE}, or \texttt{Huber}.

\subsection{Experimental Procedure and Results}
The experiments are organized into four parts. In Section~\ref{sec:demo}, we visually demonstrate the effectiveness of the proposed approach in giving unbiased prediction. In Section~\ref{sec:diffsize}, we investigate the robustness of our validation set-based tuning of the hyperparameters. In Section~\ref{sec:performance}, we intensively and quantitatively evaluate the predictive error of the proposed method and baselines with five real-world regression tasks. In Section~\ref{sec:usecase}, we demonstrate the practical benefit of our approach in a real healthcare use case, which forms the motivation for this work. See Appendix~\ref{ap_comp}-\ref{ap_ex3} for details of the experimental settings.

\subsubsection{Demonstration of Unbiased Learning}
\label{sec:demo}
\paragraph{Procedure.}
We start by conducting the experiments with synthetic data to show the effectiveness of our method in obtaining unbiased learning results from asymmetrically corrupted data with different proportions of incomplete observations, $K = \{ 25, 50, 75 \}\%$.
We use three synthetic tasks, \textbf{LowNoise}, \textbf{HighNoise}, and \textbf{Breathing}, collected from the Kaggle dataset~\citep{SagarSen2016}.
We compare the proposed method against \texttt{MSE}, which assumes that both upper- and lower-side data are correctly labeled. This comparison clarifies whether our method can learn from asymmetrically corrupted data in an unbiased manner, which \texttt{MSE} cannot do. We conducted $5$-fold cross-validation, each with a different randomly sampled training-testing split. For evaluation purposes, we do not include incomplete observations in these test sets. For each fold of the cross-validation, we use a randomly sampled $20\%$ of the training set as a validation set to choose the best hyperparameters for each algorithm.

\paragraph{Results.}
Figure~\ref{FigResultsSD} plots the error in prediction (i.e., the predicted value minus the true value) given by the proposed method and \texttt{MSE} for each data point of the three tasks with $K = 50 \%$. Since \texttt{MSE} regards both upper- and lower-side data as correctly labeled, it produces biased results due to the asymmetric label corruption, where the average error (green dashed line) is negative, which means the estimation has a negative bias. In contrast, the average error by the proposed method (blue solid line) is approximately zero. This shows that the proposed method obtained unbiased learning results. The figures for the other settings and the tables of quantitative results are provided in Appendix~\ref{ap_ex1}.

\subsubsection{Performance over different sizes of validation set}
\label{sec:diffsize}
\paragraph{Procedure.}
To demonstrate the robustness of our validation set-based approach for estimating the hyperparameters, including $\pi_{\mathrm{up}}$, we report the performance of the proposed method over different sizes of validation set. This analysis is conducted on the same tasks used in Section~\ref{sec:demo}, namely, \textbf{LowNoise}, \textbf{HighNoise}, and \textbf{Breathing}, with $K=50 \%$.

\begin{figure}[t]
	\centering
	\includegraphics[width=80mm]{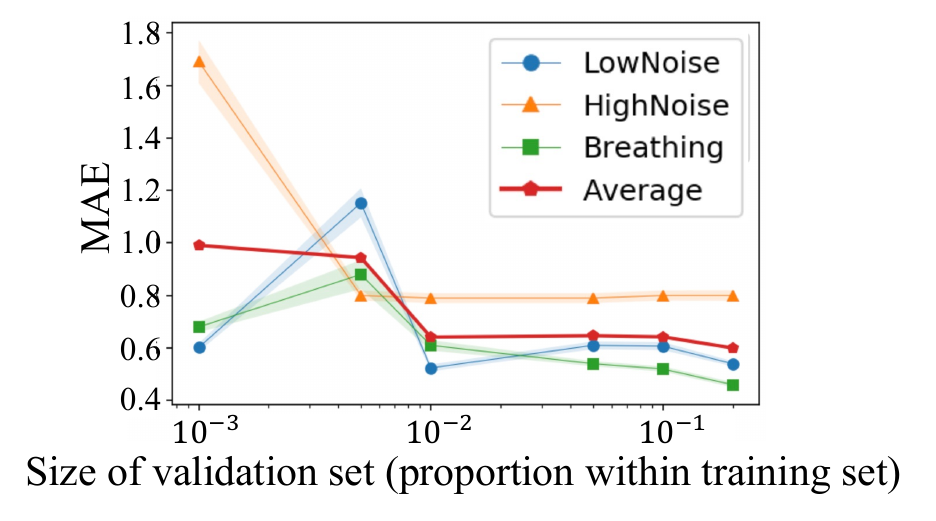}
	\caption{Performance (MAE, lower is better) of proposed method over different sizes of validation set. Blue, orange, and green lines represent the results on \textbf{LowNoise}, \textbf{HighNoise}, and \textbf{Breathing}, respectively; red line is their average. Shaded areas are confidence intervals. Leftmost point shows results when we use $0.1 \%$ of training set as a validation set, and rightmost point shows those of $20 \%$, which is the setting we used in all experiments throughout this paper.}
	\label{FigResultsDiff_valsize}
\end{figure}

\begin{figure}[t]
	\centering
	\includegraphics[width=60mm]{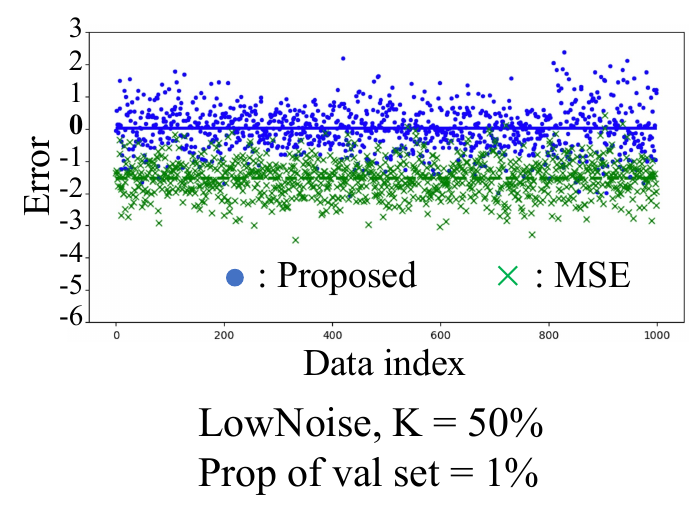}
	\caption{Errors in prediction for \textbf{LowNoise} task with $K=50 \%$ when we choose hyperparameters with $1 \%$ of the training set as a validation set. Other configurations are the same as those in Fig.~\ref{FigResultsSD}.}
	\label{FigResults1pval_Low}
\end{figure}

\paragraph{Results.}
The results are shown in Fig.~\ref{FigResultsDiff_valsize}, where we can see that the performance of the proposed method does not degrade much even when we use only $1 \%$ of the training set as the validation set. This demonstrates that the proposed approach is robust enough for small validation sets as well as for a high proportion of incomplete validation samples, as we saw in Section~\ref{sec:demo} (e.g., $K = 50\%$ and $75\%$).
Figure~\ref{FigResults1pval_Low} shows a chart similar to the ones in Fig.~\ref{FigResultsSD} (the error in prediction for the \textbf{LowNoise} task with $K=50 \%$), where we used $1 \%$ of the training set as the validation set. We can see that even in this case, the proposed method achieved approximately unbiased learning (the average error shown by the blue solid line is approximately zero). The figures for the other tasks are provided in Appendix~\ref{ap_ex1}.

\subsubsection{Performance Comparison Among Different Loss Functions}
\label{sec:performance}
\paragraph{Procedure.}
We next apply the proposed method and baselines to five different real-world healthcare tasks from the UCI Machine Learning Repository~\citep{Velloso2013,velloso2013qualitative} to provide a more extensive comparison between the proposed method and the baselines (\texttt{MSE}, \texttt{MAE}, and \texttt{Huber}). For the proposed method, we use two implementations of $L(f(\bx),y)$ for $f(\bx)\leq y'$ in Eq.~\eqref{empiricalgradient}: the absolute loss ({\tt Proposed-1}) and the squared loss ({\tt Proposed-2}). In both implementations, we use the absolute loss, which satisfies Condition~\ref{condition_L}, for $L(f(\bx),y)$ when $y'<f(\bx)$.
Here, we report the \emph{mean absolute error} (MAE), and its standard error, of the predictions $\hat{\by}= \{\hy_n\}_{n=1}^{N}$ against the corresponding true labels $\by$ across $5$-fold cross-validation, each with a different randomly sampled training-testing split. MAE is a common metric used in the healthcare domain~\citep{lee2016physiological,yeung2002reverse,wang2005robust,srinivas2020prediction} and is defined as
$
    \mathrm{MAE}(\hby,\by) \equiv \nicefrac{1}{N}\sum^N_{n=1}|\hy_n-y_n |
$.
For each fold of the cross-validation, we use a randomly sampled $20\%$ of the training set as a validation set to choose the best hyperparameters for each algorithm, in which hyperparameters providing the highest MAE in the validation set are chosen.

\paragraph{Results.}
As shown in Table~\ref{ResultRD}, \texttt{Proposed-1} and \texttt{Proposed-2} significantly outperformed the baselines. The robust regression methods (\texttt{MAE} and \texttt{Huber}) did not improve in performance against \texttt{MSE}. \texttt{Proposed-1} and \texttt{Proposed-2} respectively reduced the MAE by more than $20\%$ and $30\%$ on average, compared with the baselines.
\begin{table*}[t]
\caption{Comparison between proposed method and baselines in terms of MAE (smaller is better). Best methods are in bold. Confidence intervals are standard errors.}
\label{ResultRD}
\centering
\begin{tabular}{cccccccc}
\toprule
& Specification & Throwing A & Lifting & Lowering & Throwing B & Avg.\\
\midrule
\texttt{MSE}& $2.38 \pm 0.03$& $1.54  \pm  0.01$& $1.42  \pm  0.01$&$1.37  \pm  0.01$& $1.21  \pm  0.01$& $1.58$\\
\texttt{MAE}& $2.14  \pm  0.02$& $1.46  \pm  0.01$& $1.44  \pm  0.01$&$1.33  \pm  0.01$& $1.31  \pm  0.01$& $1.54$\\
\texttt{Huber}& $2.04  \pm  0.02$& $1.66  \pm  0.01$& $1.45  \pm  0.01$&$1.50  \pm  0.01$& $1.32  \pm  0.01$& $1.59$\\
\texttt{Proposed-1}& $1.55  \pm  0.02$& $1.18  \pm  0.01$& $1.11  \pm  0.01$&$1.14  \pm  0.01$& $1.03  \pm  0.01$& $1.20$\\
\texttt{Proposed-2}& $\bm{1.32  \pm  0.01}$& $\bm{0.99  \pm  0.01}$& $\bm{0.94  \pm  0.01}$&$\bm{0.86  \pm  0.01}$& $\bm{0.97  \pm  0.01}$& $\bm{1.02}$\\
\bottomrule
\end{tabular}
\end{table*}
\begin{table}[t]
\caption{Proportion of correct prediction period and rate of false prediction in real use case for healthcare. We estimate intrusive sensor output from outputs of non-intrusive sensors.}
\centering
\begin{tabular}{cccc}
\toprule
Proportion of correct prediction period & $0.89$\\
Rate of false prediction& $0.016$\\
\bottomrule
\end{tabular}
\label{tabCaseResults}
\end{table}

\subsubsection{Real Use Case for Healthcare}
\label{sec:usecase}
\paragraph{Procedure.}
Finally, we demonstrate the practical benefits of our approach in a real use case in healthcare. Here, from non-intrusive bed sensors installed under each of the four legs of a bed, we estimate the motion intensity of a subject that is measured accurately but intrusively with ActiGraph, a sensor wrapped around the wrist~\citep{tryon2013activity,mullaney1980wrist,webster1982activity,cole1992automatic}.
If we can mimic the outputs from ActiGraph with outputs from the bed sensors, we can measure the motion with high accuracy and high coverage, while also easing the burden on the subject.
We evaluate the results with 3-fold cross-validation, where we sequentially consider data from one subject as a test set and the others as a training set. We use evaluation metrics designed for sleep-wake discrimination~\citep{cole1992automatic}, i.e., the proportion of correct prediction period and rate of false prediction.

\paragraph{Results.}
Table~\ref{tabCaseResults} shows the proportion of correct prediction period and rate of false prediction, where we can see that the proposed method captured $89$ percent of the total time period of the motions that were captured by ActiGraph, and false detection due to factors such as floor vibration was only $1.6$ percent. Furthermore, the proposed method captured $15$ additional motions that were not captured by ActiGraph. The baseline method \texttt{MSE} was severely underfitted, and most of the weights were zero; thus, we omitted these results.
Overall, our findings here demonstrate that ActiGraph can be replaced with bed sensors, and we can also use the bed sensors for the inputs of certain ActiGraph functions, such as sleep-wake discrimination~\citep{cole1992automatic}. See Appendix~\ref{ap_ex3} for further details, including the actual estimation results of the motion intensity.

\section{Discussion}
\paragraph{Limitations.}
In this paper, we do not address ordinary outliers, where the coverage and incompleteness are consistent between a label and explanatory variables.
Other established approaches can handle such cases.
Only when the corruption is asymmetric does it lead to the technical challenge we address here.
In that sense, we can handle the opposite asymmetric corruption, in which labels for some observations may become inconsistently \emph{higher} than those for typical observations. This can be handled by a learning algorithm from \emph{lower-side labeled data and unlabeled data}, i.e., LU regression.
Since our derivation of U2 regression is straightforwardly applicable to this LU regression case, we show only its learning algorithm in Appendix~\ref{ap_LU}.
Such a scenario can be found when a sensor for the label has ideal coverage but sensors for explanatory variables have smaller coverage. This may correspond to more challenging sensor replacement, which addresses critical invasiveness, cost, and lack of alternative methods.
In that case, there are unidentifiable incomplete observations in explanatory variables, which leads to the opposite asymmetric data corruption.

\paragraph{Related problems in classification.}
In the classification problem setting, positive-unlabeled (PU) learning addresses a problem related to our problem setting, where it is assumed that negative data cannot be obtained, but unlabeled data are available as well as positive data~\citep{denis1998pac,de1999positive,letouzey2000learning,shi2018positive,kato2018learning,sakai2019covariate,PUPG_Li_2019,zhang2019positive,zhanglearning2020,chen2020SelfPU,chen2020variational,luo2021pulns,hu2021predictive,li2021your,wilton2022positive}. An unbiased risk estimator has also been proposed~\citep{du2014analysis,du2015convex}. However, PU learning cannot be used for a regression problem, where labels are real values and we need to handle order and gradation between labels. Also, these studies do not address the asymmetric noise we examine in this paper, as their motivation is rather to address the labeling cost of negative labels.

\paragraph{Other possible use cases.}
Examples of predicting the magnitude values of a sensor, which is a field of application of U2 regression, can be found in several areas.
One example is estimating the wind speed or rainfall in a specific region from observable macroscopic information~\citep{cheng2008semi,abraham2010integrated,abraham2013distribution,vandal2017deepsd}, known as statistical downscaling~\citep{wilby2004guidelines}. Wind speed and rainfall, which are labels in these tasks, can be sensed locally in a limited number of locations and provide incomplete observations and biased labels compared with macroscopic information, which is considered to be explanatory variables.
Molecular biology is another interesting example~\cite{seal2020estimating,dizaji2021deep}, where a low measurement of a gene expression may indicate that the gene is not present or that the sensor failed to detect it.

\paragraph{Future work.}
We showed that our approach to estimating hyperparameters based on the grid search with the validation set was effective even for the important ratio for upper-side labeled data, $p(f_t(\bx)\leq y)$. It also provides the flexibility needed to handle data variation. Most studies on PU learning assume that a hyperparameter corresponding to $\pi_{\mathrm{up}}$ is given~\citep{hammoudeh2020learning,sonntag2021predicting,lin2022federated}, and some papers have addressed this hyperparameter estimation as their main contribution~\citep{jain2016estimating,ramaswamy2016mixture,christoffel2016class,jain2020class}.
Developing a method for the hyperparameter estimation to improve performance would be a worthwhile next step of our own study. Also, in Assumption~\ref{assumption_obsprocess}, we assumed $\epsilon_{\mathrm{s}} \indep f^*(\bx)$ and $\epsilon_{\mathrm{a}} \indep f^*(\bx)$, which is a common noise assumption. Addressing the case where the noises are not independent of $f^*(\bx)$ is another possible future direction of our work.

\paragraph{Conclusion.}
We formulated a regression problem from asymmetrically corrupted data in which training data are corrupted with an asymmetric noise that always has a negative value. This causes labels for data with relatively lower label values to be particularly unreliable.
To address this problem, we proposed a learning algorithm called U2 regression. Under some technical assumptions, we showed that our algorithm is unbiased and consistent with the one that uses uncorrupted data. Our analysis is based on the equivalence of the gradient between them.
An experimental evaluation demonstrated that the proposed method performed significantly better than the methods without the assumption of the asymmetrical label corruption.

\bibliography{ref.bib}

\begin{thebibliography}{67}
\providecommand{\natexlab}[1]{#1}
\providecommand{\url}[1]{\texttt{#1}}
\expandafter\ifx\csname urlstyle\endcsname\relax
  \providecommand{\doi}[1]{doi: #1}\else
  \providecommand{\doi}{doi: \begingroup \urlstyle{rm}\Url}\fi

\bibitem[Abraham \& Tan(2010)Abraham and Tan]{abraham2010integrated}
Abraham, Z. and Tan, P.-N.
\newblock An integrated framework for simultaneous classification and regression of time-series data.
\newblock In \emph{SDM}, pp.\  653--664, 2010.

\bibitem[Abraham et~al.(2013)Abraham, Tan, Perdinan, Winkler, Zhong, and Liszewska]{abraham2013distribution}
Abraham, Z., Tan, P.-N., Perdinan, Winkler, J.~A., Zhong, S., and Liszewska, M.
\newblock Distribution regularized regression framework for climate modeling.
\newblock In \emph{SDM}, pp.\  333--341, 2013.

\bibitem[Alaziz et~al.(2016)Alaziz, Jia, Liu, Howard, Chen, and Zhang]{alaziz2016motion}
Alaziz, M., Jia, Z., Liu, J., Howard, R., Chen, Y., and Zhang, Y.
\newblock Motion scale: A body motion monitoring system using bed-mounted wireless load cells.
\newblock In \emph{CHASE}, pp.\  183--192, 2016.

\bibitem[Alaziz et~al.(2017)Alaziz, Jia, Howard, Lin, and Zhang]{alaziz2017motiontree}
Alaziz, M., Jia, Z., Howard, R., Lin, X., and Zhang, Y.
\newblock Motiontree: a tree-based in-bed body motion classification system using load-cells.
\newblock In \emph{CHASE}, pp.\  127--136, 2017.

\bibitem[Carlson et~al.(2018)Carlson, Suliman, Alivar, Prakash, Thompson, Natarajan, and Warren]{carlson2018pilot}
Carlson, C., Suliman, A., Alivar, A., Prakash, P., Thompson, D., Natarajan, B., and Warren, S.
\newblock A pilot study of an unobtrusive bed-based sleep quality monitor for severely disabled autistic children.
\newblock In \emph{EMBC}, pp.\  4343--4346, 2018.

\bibitem[Chen et~al.(2020{\natexlab{a}})Chen, Liu, Wang, Zhao, and Wu]{chen2020variational}
Chen, H., Liu, F., Wang, Y., Zhao, L., and Wu, H.
\newblock A variational approach for learning from positive and unlabeled data.
\newblock In \emph{NeurIPS}, 2020{\natexlab{a}}.

\bibitem[Chen et~al.(2020{\natexlab{b}})Chen, Chen, Chen, Yuan, Gong, Chen, and Wang]{chen2020SelfPU}
Chen, X., Chen, W., Chen, T., Yuan, Y., Gong, C., Chen, K., and Wang, Z.
\newblock Self-{PU}: Self boosted and calibrated positive-unlabeled training.
\newblock In \emph{ICML}, 2020{\natexlab{b}}.

\bibitem[Cheng \& Tan(2008)Cheng and Tan]{cheng2008semi}
Cheng, H. and Tan, P.-N.
\newblock Semi-supervised learning with data calibration for long-term time series forecasting.
\newblock In \emph{KDD}, pp.\  133--141, 2008.

\bibitem[Christoffel et~al.(2016)Christoffel, Niu, and Sugiyama]{christoffel2016class}
Christoffel, M., Niu, G., and Sugiyama, M.
\newblock Class-prior estimation for learning from positive and unlabeled data.
\newblock In \emph{ACML}, pp.\  221--236. PMLR, 2016.

\bibitem[Chung(1968)]{chung1968course}
Chung, K.~L.
\newblock \emph{A course in probability theory}.
\newblock Academic press, 1968.

\bibitem[Cole et~al.(1992)Cole, Kripke, Gruen, Mullaney, and Gillin]{cole1992automatic}
Cole, R.~J., Kripke, D.~F., Gruen, W., Mullaney, D.~J., and Gillin, J.~C.
\newblock Automatic sleep/wake identification from wrist activity.
\newblock \emph{Sleep}, 15\penalty0 (5):\penalty0 461--469, 1992.

\bibitem[De~Comit{\'e} et~al.(1999)De~Comit{\'e}, Denis, Gilleron, and Letouzey]{de1999positive}
De~Comit{\'e}, F., Denis, F., Gilleron, R., and Letouzey, F.
\newblock Positive and unlabeled examples help learning.
\newblock In \emph{ALT}, pp.\  219--230, 1999.

\bibitem[Denis(1998)]{denis1998pac}
Denis, F.
\newblock Pac learning from positive statistical queries.
\newblock In \emph{ALT}, pp.\  112--126, 1998.

\bibitem[Dizaji et~al.(2021)Dizaji, Chen, and Huang]{dizaji2021deep}
Dizaji, K.~G., Chen, W., and Huang, H.
\newblock Deep large-scale multitask learning network for gene expression inference.
\newblock \emph{Journal of Computational Biology}, 28\penalty0 (5):\penalty0 485--500, 2021.

\bibitem[Draper \& Smith(1998)Draper and Smith]{draper1998applied}
Draper, N.~R. and Smith, H.
\newblock \emph{Applied regression analysis}, volume 326.
\newblock John Wiley \& Sons, 1998.

\bibitem[Du~Plessis et~al.(2015)Du~Plessis, Niu, and Sugiyama]{du2015convex}
Du~Plessis, M., Niu, G., and Sugiyama, M.
\newblock Convex formulation for learning from positive and unlabeled data.
\newblock In \emph{ICML}, pp.\  1386--1394, 2015.

\bibitem[Du~Plessis et~al.(2014)Du~Plessis, Niu, and Sugiyama]{du2014analysis}
Du~Plessis, M.~C., Niu, G., and Sugiyama, M.
\newblock Analysis of learning from positive and unlabeled data.
\newblock In \emph{NIPS}, pp.\  703--711, 2014.

\bibitem[Enders(2010)]{enders2010applied}
Enders, C.~K.
\newblock \emph{Applied missing data analysis}.
\newblock Guilford press, 2010.

\bibitem[Hammoudeh \& Lowd(2020)Hammoudeh and Lowd]{hammoudeh2020learning}
Hammoudeh, Z. and Lowd, D.
\newblock Learning from positive and unlabeled data with arbitrary positive shift.
\newblock In \emph{NeurIPS}, 2020.

\bibitem[Hu et~al.(2021)Hu, Le, Liu, Ji, Ma, Zhao, and Yan]{hu2021predictive}
Hu, W., Le, R., Liu, B., Ji, F., Ma, J., Zhao, D., and Yan, R.
\newblock Predictive adversarial learning from positive and unlabeled data.
\newblock In \emph{AAAI}, pp.\  7806--7814, 2021.

\bibitem[Huber et~al.(1964)]{huber1964robust}
Huber, P.~J. et~al.
\newblock Robust estimation of a location parameter.
\newblock \emph{The Annals of Mathematical Statistics}, 35\penalty0 (1):\penalty0 73--101, 1964.

\bibitem[Inan et~al.(2009)Inan, Etemadi, Paloma, Giovangrandi, and Kovacs]{inan2009non}
Inan, O., Etemadi, M., Paloma, A., Giovangrandi, L., and Kovacs, G.
\newblock Non-invasive cardiac output trending during exercise recovery on a bathroom-scale-based ballistocardiograph.
\newblock \emph{Physiological measurement}, 30\penalty0 (3):\penalty0 261, 2009.

\bibitem[Jain et~al.(2016)Jain, White, and Radivojac]{jain2016estimating}
Jain, S., White, M., and Radivojac, P.
\newblock Estimating the class prior and posterior from noisy positives and unlabeled data.
\newblock In \emph{NeurIPS}, pp.\  2693--2701, 2016.

\bibitem[Jain et~al.(2020)Jain, Delano, Sharma, and Radivojac]{jain2020class}
Jain, S., Delano, J., Sharma, H., and Radivojac, P.
\newblock Class prior estimation with biased positives and unlabeled examples.
\newblock In \emph{AAAI}, pp.\  4255--4263, 2020.

\bibitem[Jean et~al.(2018)Jean, Xie, and Ermon]{jean2018semi}
Jean, N., Xie, S.~M., and Ermon, S.
\newblock Semi-supervised deep kernel learning: Regression with unlabeled data by minimizing predictive variance.
\newblock In \emph{NeurIPS}, 2018.

\bibitem[Kato et~al.(2019)Kato, Teshima, and Honda]{kato2018learning}
Kato, M., Teshima, T., and Honda, J.
\newblock Learning from positive and unlabeled data with a selection bias.
\newblock In \emph{ICLR}, 2019.

\bibitem[Kingma \& Ba(2015)Kingma and Ba]{kingma2014adam}
Kingma, D. and Ba, J.
\newblock Adam: A method for stochastic optimization.
\newblock In \emph{ICLR}, 2015.

\bibitem[Lee et~al.(2016)Lee, Yoon, Han, Joo, and Park]{lee2016physiological}
Lee, W., Yoon, H., Han, C., Joo, K., and Park, K.
\newblock Physiological signal monitoring bed for infants based on load-cell sensors.
\newblock \emph{Sensors}, 16\penalty0 (3):\penalty0 409, 2016.

\bibitem[Letouzey et~al.(2000)Letouzey, Denis, and Gilleron]{letouzey2000learning}
Letouzey, F., Denis, F., and Gilleron, R.
\newblock Learning from positive and unlabeled examples.
\newblock In \emph{ALT}, pp.\  71--85, 2000.

\bibitem[Li et~al.(2021)Li, Li, Feng, and Ouyang]{li2021your}
Li, C., Li, X., Feng, L., and Ouyang, J.
\newblock Who is your right mixup partner in positive and unlabeled learning.
\newblock In \emph{ICLR}, 2021.

\bibitem[Li et~al.(2019)Li, Wang, Ma, Ortal, Zhao, Stenger, and Hirate]{PUPG_Li_2019}
Li, T., Wang, C.-C., Ma, Y., Ortal, P., Zhao, Q., Stenger, B., and Hirate, Y.
\newblock Learning classifiers on positive and unlabeled data with policy gradient.
\newblock In \emph{ICDM}, pp.\  399--408, 2019.

\bibitem[Lin et~al.(2022)Lin, Chen, Xu, Xu, Gui, Deng, and Wang]{lin2022federated}
Lin, X., Chen, H., Xu, Y., Xu, C., Gui, X., Deng, Y., and Wang, Y.
\newblock Federated learning with positive and unlabeled data.
\newblock In \emph{ICML}, pp.\  13344--13355, 2022.

\bibitem[Luo et~al.(2021)Luo, Zhao, Chen, Qiao, Du, Zhang, Wu, Cai, He, Rajmohan, et~al.]{luo2021pulns}
Luo, C., Zhao, P., Chen, C., Qiao, B., Du, C., Zhang, H., Wu, W., Cai, S., He, B., Rajmohan, S., et~al.
\newblock Pulns: Positive-unlabeled learning with effective negative sample selector.
\newblock In \emph{AAAI}, pp.\  8784--8792, 2021.

\bibitem[Ma \& Chen(2019)Ma and Chen]{ma2019missing}
Ma, W. and Chen, G.~H.
\newblock Missing not at random in matrix completion: The effectiveness of estimating missingness probabilities under a low nuclear norm assumption.
\newblock In \emph{NeurIPS}, 2019.

\bibitem[Mullaney et~al.(1980)Mullaney, Kripke, and Messin]{mullaney1980wrist}
Mullaney, D., Kripke, D., and Messin, S.
\newblock Wrist-actigraphic estimation of sleep time.
\newblock \emph{Sleep}, 3\penalty0 (1):\penalty0 83--92, 1980.

\bibitem[Nair \& Hinton(2010)Nair and Hinton]{nair2010rectified}
Nair, V. and Hinton, G.~E.
\newblock Rectified linear units improve restricted boltzmann machines.
\newblock In \emph{ICML}, pp.\  807--814, 2010.

\bibitem[Narula \& Wellington(1982)Narula and Wellington]{narula1982minimum}
Narula, S.~C. and Wellington, J.~F.
\newblock The minimum sum of absolute errors regression: A state of the art survey.
\newblock \emph{International Statistical Review/Revue Internationale de Statistique}, pp.\  317--326, 1982.

\bibitem[Nukaya et~al.(2010)Nukaya, Shino, Kurihara, Watanabe, and Tanaka]{nukaya2010noninvasive}
Nukaya, S., Shino, T., Kurihara, Y., Watanabe, K., and Tanaka, H.
\newblock Noninvasive bed sensing of human biosignals via piezoceramic devices sandwiched between the floor and bed.
\newblock \emph{IEEE Sensors journal}, 12\penalty0 (3):\penalty0 431--438, 2010.

\bibitem[Pearson(2006)]{pearson2006problem}
Pearson, R.~K.
\newblock The problem of disguised missing data.
\newblock \emph{Acm Sigkdd Explorations Newsletter}, 8\penalty0 (1):\penalty0 83--92, 2006.

\bibitem[Qahtan et~al.(2018)Qahtan, Elmagarmid, Castro~Fernandez, Ouzzani, and Tang]{qahtan2018fahes}
Qahtan, A.~A., Elmagarmid, A., Castro~Fernandez, R., Ouzzani, M., and Tang, N.
\newblock Fahes: A robust disguised missing values detector.
\newblock In \emph{KDD}, pp.\  2100--2109, 2018.

\bibitem[Ramaswamy et~al.(2016)Ramaswamy, Scott, and Tewari]{ramaswamy2016mixture}
Ramaswamy, H., Scott, C., and Tewari, A.
\newblock Mixture proportion estimation via kernel embeddings of distributions.
\newblock In \emph{ICML}, pp.\  2052--2060. PMLR, 2016.

\bibitem[Sakai \& Shimizu(2019)Sakai and Shimizu]{sakai2019covariate}
Sakai, T. and Shimizu, N.
\newblock Covariate shift adaptation on learning from positive and unlabeled data.
\newblock In \emph{AAAI}, pp.\  4838--4845, 2019.

\bibitem[Seal et~al.(2020)Seal, Das, Goswami, and De]{seal2020estimating}
Seal, D.~B., Das, V., Goswami, S., and De, R.~K.
\newblock Estimating gene expression from dna methylation and copy number variation: a deep learning regression model for multi-omics integration.
\newblock \emph{Genomics}, 112\penalty0 (4):\penalty0 2833--2841, 2020.

\bibitem[Sen(2016)]{SagarSen2016}
Sen, S.
\newblock Kaggle dataset.
\newblock \url{https://www.kaggle.com/sagarsen/breathing-data-from-a-chest-belt}, 2016.

\bibitem[Shi et~al.(2018)Shi, Pan, Yang, and Gong]{shi2018positive}
Shi, H., Pan, S., Yang, J., and Gong, C.
\newblock Positive and unlabeled learning via loss decomposition and centroid estimation.
\newblock In \emph{IJCAI}, pp.\  2689--2695, 2018.

\bibitem[Shi et~al.(2017)Shi, Gao, Lausen, Wang, Yeung, Wong, and Woo]{shi2017deep}
Shi, X., Gao, Z., Lausen, L., Wang, H., Yeung, D.-Y., Wong, W.-k., and Woo, W.-c.
\newblock Deep learning for precipitation nowcasting: A benchmark and a new model.
\newblock In \emph{NIPS}, pp.\  5617--5627, 2017.

\bibitem[Smieja et~al.(2018)Smieja, Struski, Tabor, Zieli{\'n}ski, and Spurek]{smieja2018processing}
Smieja, M., Struski, {\L}., Tabor, J., Zieli{\'n}ski, B., and Spurek, P.
\newblock Processing of missing data by neural networks.
\newblock In \emph{NeurIPS}, 2018.

\bibitem[Sonntag et~al.(2021)Sonntag, Engel, and Schmidt-Thieme]{sonntag2021predicting}
Sonntag, J., Engel, M., and Schmidt-Thieme, L.
\newblock Predicting parking availability from mobile payment transactions with positive unlabeled learning.
\newblock In \emph{AAAI}, pp.\  15408--15415, 2021.

\bibitem[Sportisse et~al.(2020)Sportisse, Boyer, and Josse]{sportisse2020estimation}
Sportisse, A., Boyer, C., and Josse, J.
\newblock Estimation and imputation in probabilistic principal component analysis with missing not at random data.
\newblock In \emph{NeurIPS}, 2020.

\bibitem[Srinivas et~al.(2020)Srinivas, Rani, Rao, Patra, Madhukar, and Mahendar]{srinivas2020prediction}
Srinivas, K., Rani, B.~K., Rao, M., Patra, R.~K., Madhukar, G., and Mahendar, A.
\newblock Prediction of heart disease using hybrid linear regression.
\newblock \emph{European Journal of Molecular \& Clinical Medicine}, 7\penalty0 (5):\penalty0 1159--1171, 2020.

\bibitem[Srivastava et~al.(2014)Srivastava, Hinton, Krizhevsky, Sutskever, and Salakhutdinov]{srivastava2014dropout}
Srivastava, N., Hinton, G., Krizhevsky, A., Sutskever, I., and Salakhutdinov, R.
\newblock Dropout: a simple way to prevent neural networks from overfitting.
\newblock \emph{The journal of machine learning research}, 15\penalty0 (1):\penalty0 1929--1958, 2014.

\bibitem[Tanaka et~al.(2019)Tanaka, Iwata, Tanaka, Kurashima, Okawa, and Toda]{tanaka2019refining}
Tanaka, Y., Iwata, T., Tanaka, T., Kurashima, T., Okawa, M., and Toda, H.
\newblock Refining coarse-grained spatial data using auxiliary spatial data sets with various granularities.
\newblock In \emph{AAAI}, volume~33, pp.\  5091--5099, 2019.

\bibitem[Tryon(2013)]{tryon2013activity}
Tryon, W.~W.
\newblock \emph{Activity measurement in psychology and medicine}.
\newblock Springer Science \& Business Media, 2013.

\bibitem[Vandal et~al.(2017)Vandal, Kodra, Ganguly, Michaelis, Nemani, and Ganguly]{vandal2017deepsd}
Vandal, T., Kodra, E., Ganguly, S., Michaelis, A., Nemani, R., and Ganguly, A.~R.
\newblock Deepsd: Generating high resolution climate change projections through single image super-resolution.
\newblock In \emph{KDD}, pp.\  1663--1672, 2017.

\bibitem[Velloso(2013)]{Velloso2013}
Velloso, E.
\newblock {UCI} machine learning repository.
\newblock \url{https://archive.ics.uci.edu/ml/datasets/Weight+Lifting+Exercises+monitored+with+Inertial+Measurement+Units}, 2013.

\bibitem[Velloso et~al.(2013)Velloso, Bulling, Gellersen, Ugulino, and Fuks]{velloso2013qualitative}
Velloso, E., Bulling, A., Gellersen, H., Ugulino, W., and Fuks, H.
\newblock Qualitative activity recognition of weight lifting exercises.
\newblock In \emph{AH}, pp.\  116--123, 2013.

\bibitem[Wang et~al.(2005)Wang, He, and Chen]{wang2005robust}
Wang, Z., He, Z., and Chen, J.~D.
\newblock Robust time delay estimation of bioelectric signals using least absolute deviation neural network.
\newblock \emph{IEEE Transactions on biomedical engineering}, 52\penalty0 (3):\penalty0 454--462, 2005.

\bibitem[Webster et~al.(1982)Webster, Kripke, Messin, Mullaney, and Wyborney]{webster1982activity}
Webster, J.~B., Kripke, D.~F., Messin, S., Mullaney, D.~J., and Wyborney, G.
\newblock An activity-based sleep monitor system for ambulatory use.
\newblock \emph{Sleep}, 5\penalty0 (4):\penalty0 389--399, 1982.

\bibitem[Wilby et~al.(2004)Wilby, Charles, Zorita, Timbal, Whetton, and Mearns]{wilby2004guidelines}
Wilby, R.~L., Charles, S., Zorita, E., Timbal, B., Whetton, P., and Mearns, L.
\newblock Guidelines for use of climate scenarios developed from statistical downscaling methods.
\newblock \emph{Supporting material of the Intergovernmental Panel on Climate Change, available from the DDC of IPCC TGCIA}, 27, 2004.

\bibitem[Wilcox(1997)]{wilcox2012introduction}
Wilcox, R.~R.
\newblock \emph{Introduction to robust estimation and hypothesis testing}.
\newblock Academic Press, 1997.

\bibitem[Wilton et~al.(2022)Wilton, Koay, Ko, Xu, and Ye]{wilton2022positive}
Wilton, J., Koay, A., Ko, R., Xu, M., and Ye, N.
\newblock Positive-unlabeled learning using random forests via recursive greedy risk minimization.
\newblock In \emph{NeurIPS}, 2022.

\bibitem[Yeung et~al.(2002)Yeung, Tegn{\'e}r, and Collins]{yeung2002reverse}
Yeung, M.~S., Tegn{\'e}r, J., and Collins, J.~J.
\newblock Reverse engineering gene networks using singular value decomposition and robust regression.
\newblock \emph{Proceedings of the National Academy of Sciences}, 99\penalty0 (9):\penalty0 6163--6168, 2002.

\bibitem[Zhang et~al.(2019)Zhang, Ren, Liu, Yang, and Gong]{zhang2019positive}
Zhang, C., Ren, D., Liu, T., Yang, J., and Gong, C.
\newblock Positive and unlabeled learning with label disambiguation.
\newblock In \emph{IJCAI}, pp.\  1--7, 2019.

\bibitem[Zhang et~al.(2020)Zhang, Hou, and Zhang]{zhanglearning2020}
Zhang, C., Hou, Y., and Zhang, Y.
\newblock Learning from positive and unlabeled data without explicit estimation of class prior.
\newblock In \emph{AAAI}, pp.\  6762--6769, 2020.

\bibitem[Zhou et~al.(2019)Zhou, Li, Zhou, Zhu, and Ye]{zhou2019graph}
Zhou, F., Li, T., Zhou, H., Zhu, H., and Ye, J.
\newblock Graph-based semi-supervised learning with non-ignorable non-response.
\newblock In \emph{NeurIPS}, 2019.

\bibitem[Zhou \& Li(2005)Zhou and Li]{zhou2005semi}
Zhou, Z.-H. and Li, M.
\newblock Semi-supervised regression with co-training.
\newblock In \emph{IJCAI}, pp.\  908--913, 2005.

\bibitem[Zhu \& Goldberg(2009)Zhu and Goldberg]{zhu2009introduction}
Zhu, X. and Goldberg, A.~B.
\newblock Introduction to semi-supervised learning.
\newblock \emph{Synthesis lectures on artificial intelligence and machine learning}, 3\penalty0 (1):\penalty0 1--130, 2009.

\end{thebibliography}
\bibliographystyle{icml2023}


\appendix
\section{Proofs}
\subsection{Proof of Lemma~\ref{lemma_obsprocess_E}}
\label{ap_pr_lemma_obsprocess_E}
\begin{proof}
For the proof of Lemma~\ref{lemma_obsprocess_E}, we will derive two important lemmas from Assumption~\ref{assumption_obsprocess}. Then, we will prove Lemma~\ref{lemma_obsprocess_E} by using them.

We first show $f^*(\bx)\leq y' \Rightarrow \epsilon_{\mathrm{a}}=0$.
When $f^*(\bx)\leq y'$, we have the following from Eqs.~\eqref{observation_process_unbiased} and~\eqref{observation_process}:
\begin{align}
    f^*(\bx) &\leq f^*(\bx) + \epsilon_{\mathrm{s}} + \epsilon_{\mathrm{a}}\\
    0 &\leq \epsilon_{\mathrm{s}} + \epsilon_{\mathrm{a}}\nonumber\\
    -\epsilon_{\mathrm{a}} &\leq \epsilon_{\mathrm{s}}.\nonumber
\end{align}
Since $\epsilon_{\mathrm{a}} \leq 0$ by Assumption~\ref{assumption_obsprocess}, we have
\begin{align}
    \label{ineq_lo}
    |\epsilon_{\mathrm{a}}| \leq \epsilon_{\mathrm{s}}.
\end{align}
If $\epsilon_{\mathrm{a}}<0$, Assumption~\ref{assumption_obsprocess} implies $|\epsilon_{\mathrm{s}}|<|\epsilon_{\mathrm{a}}|$, which contradicts Eq.~\eqref{ineq_lo}. Hence, we must have
\begin{align}
    \epsilon_{\mathrm{a}} =0.
\end{align}

Since $y=y'$ when $\epsilon_{\mathrm{a}}=0$, we have
\begin{align}
    \label{derivation_of_a0lemma}
    p(\bx,y'|f^*(\bx)\leq y')
    & = p(\bx,y'|f^*(\bx)\leq y', \epsilon_{\mathrm{a}}=0)\nonumber\\
    & = p(\bx,y|f^*(\bx)\leq y, \epsilon_{\mathrm{a}}=0)\nonumber\\
    & = p(\bx,y|f^*(\bx)\leq y),
\end{align}
which establishes
\begin{lemma}
    \label{lemma_obsprocess}
    Let $p(\bx,y,y')$ be the underlying probability distribution for $\bx$, $y$, and $y'$. Then,
  \begin{align}
      \label{consistent_upper}
      p(\bx,y'|f^*(\bx)\leq y') &= p(\bx,y|f^*(\bx)\leq y).
  \end{align}
\end{lemma}

Similar to Lemma~\ref{lemma_obsprocess}, we show $f(x)\leq y'\Rightarrow\epsilon_{\mathrm{a}}=0$.
Let $\calF' \equiv \{f\in\calF : |f(\bx)-f^*(\bx)| \leq |\epsilon_{\mathrm{s}}| \mbox{ a.s.}\}$, which represents our natural expectation that the regression function $f$ well approximates $f^*$.
When $f(\bx)\leq y'$, we have the following from Eqs.~\eqref{observation_process_unbiased} and~\eqref{observation_process} with the condition $f\in\calF'$:
\begin{align}
    f(\bx) &\leq f^*(\bx) + \epsilon_{\mathrm{s}} + \epsilon_{\mathrm{a}}\\
    f(\bx) &\leq f(\bx) + \epsilon_{\mathrm{s}} + \epsilon_{\mathrm{a}}+|\epsilon_{\mathrm{s}}|\nonumber\\
    0 &\leq \epsilon_{\mathrm{s}} + \epsilon_{\mathrm{a}}+|\epsilon_{\mathrm{s}}|\nonumber\\
    -\epsilon_{\mathrm{a}} &\leq \epsilon_{\mathrm{s}}+|\epsilon_{\mathrm{s}}|.\nonumber
\end{align}
Since $\epsilon_{\mathrm{a}} \leq 0$ by Assumption~\ref{assumption_obsprocess}, we have
\begin{align}
    \label{ineq_lo2}
    |\epsilon_{\mathrm{a}}| \leq \epsilon_{\mathrm{s}}+|\epsilon_{\mathrm{s}}|.
\end{align}
If $\epsilon_{\mathrm{a}}<0$, Assumption~\ref{assumption_obsprocess} implies $2|\epsilon_{\mathrm{s}}| < |\epsilon_{\mathrm{a}}|$, which contradicts Eq.~\eqref{ineq_lo2}. Hence, we must have
\begin{align}
    \epsilon_{\mathrm{a}} =0.
\end{align}
Since $y=y'$ when $\epsilon_{\mathrm{a}}=0$, by replacing $f^*$ with $f$ for the argument in the derivation of Lemma~\ref{lemma_obsprocess} in Eq.~\eqref{derivation_of_a0lemma}, we have
\begin{lemma}
    \label{lemma_f}
    Let $\calF' \equiv \{f\in\calF : |f(\bx)-f^*(\bx)| \leq |\epsilon_{\mathrm{s}}|\}$. When $f\in\calF'$, the following holds:
    \begin{align}
        \label{consistent_upper2}
        p(x,y'|f(\bx)\leq y') &= p(x,y|f(\bx)\leq y).
    \end{align}
\end{lemma}

Lemma~\ref{lemma_obsprocess} immediately implies
\begin{align}
  \label{consistent_E_upper}
  \mathbb{E}_{p(\bx,y'|f^*(\bx)\leq y')}[G(\bx,y')] &= \mathbb{E}_{p(\bx,y|f^*(\bx)\leq y)}[G(\bx,y)]
\end{align}
for any function $G:\mathbb{R}^D\times\mathbb{R}\to\mathbb{R}$ as long as the expectations exist.
When $f\in\calF'$, from Lemma~\ref{lemma_f}, we then have
\begin{align}
    \label{consistent_E_upper2A}
    \mathbb{E}_{p(\bx,y'|f(\bx)\leq y')}[G(\bx,y')] &= \mathbb{E}_{p(\bx,y|f(\bx)\leq y)}[G(\bx,y)].
\end{align}

\end{proof}

\subsection{Proof of Proposition~\ref{proposition1}}
\label{ap_pr_proposition1}
\begin{proof}
From the decomposed gradient $\nabla \mathcal{L}(f_t)$ in Eq.~\eqref{decomposedGradient_new}, we derive the proposed gradient only with the expectations over $p(\bx,y')$.

From Condition~\ref{condition_L} for $L(f(\bx),y)$, $\nabla L(f(\bx),y) = \bg(f(\bx))$ when $y<f(\bx)$.
Thus, Eq.~\eqref{decomposedGradient_new} can be rewritten as
\begin{align}
    \label{decomposedGradient_with_Ldash}
    \nabla \mathcal{L}(f_t) =&p(f_t(\bx)\leq y) \mathbb{E}_{p(\bx,y|f_t(\bx)\leq y)}[\nabla L(f_t(\bx),y)]\\\nonumber
    &+ p(y<f_t(\bx)) \mathbb{E}_{p(\bx|y<f_t(\bx))}\Big[\bg(f_t(\bx))\Big],
\end{align}
where $y$ is marginalized out in the expectation in the second term since $\bg(f_t(\bx))$ does not depend on $y$.

Here, Eqs.~\eqref{gradient_new} and~\eqref{decomposedGradient_new} can be rewritten by replacing $\nabla L(f_t(\bx),y)$ with $\bg(f_t(\bx))$, as
\begin{align}
    &\mathbb{E}_{p(\bx,y)}[\bg(f_t(\bx))]\\\nonumber
    &~~~~~~~~~~~~~~~~~~~~=p(f_t(\bx)\leq y) \mathbb{E}_{p(\bx,y|f_t(\bx)\leq y)}\Big[\bg(f_t(\bx))\Big]\\\nonumber
    &~~~~~~~~~~~~~~~~~~~~~~~~+ p(y<f_t(\bx)) \mathbb{E}_{p(\bx,y|y<f_t(\bx))}\Big[\bg(f_t(\bx))\Big]\nonumber\\
\label{unlabeledGradient2_new}
    &p(y<f_t(\bx)) \mathbb{E}_{p(\bx,y|y<f_t(\bx))}\Big[\bg(f_t(\bx))\Big]\\\nonumber
    &~~~~~~~~~~~~~~~~~~=\mathbb{E}_{p(\bx,y)}\Big[\bg(f_t(\bx))\Big]\\\nonumber
    &~~~~~~~~~~~~~~~~~~~~~~- p(f_t(\bx)\leq y) \mathbb{E}_{p(\bx,y|f_t(\bx)\leq y)}\Big[\bg(f_t(\bx))\Big].
\end{align}
Since $\bg(f_t(\bx))$ does not depend on $y$, we can marginalize out $y$ in Eq.~\eqref{unlabeledGradient2_new} as
\begin{align}
    \label{unlabeledGradient2_new2}
    &p(y<f_t(\bx)) \mathbb{E}_{p(\bx|y<f_t(\bx))}\Big[\bg(f_t(\bx))\Big]\\\nonumber
    &~~~~~~~~~~~~~~~~~~=\mathbb{E}_{p(\bx)}\Big[\bg(f_t(\bx))\Big]\\\nonumber
    &~~~~~~~~~~~~~~~~~~~~~~- p(f_t(\bx)\leq y) \mathbb{E}_{p(\bx|f_t(\bx)\leq y)}\Big[\bg(f_t(\bx))\Big].
\end{align}

From Eq.~\eqref{unlabeledGradient2_new2}, we can express Eq.~\eqref{decomposedGradient_with_Ldash} as
\begin{align}
    \label{replacedGradient_new}
    \nabla \mathcal{L}(f_t) =&p(f_t(\bx)\leq y) \mathbb{E}_{p(\bx,y|f_t(\bx)\leq y)}[\nabla L(f_t(\bx),y)]\nonumber\\
    &+ \mathbb{E}_{p(\bx)}\Big[\bg(f_t(\bx))\Big]\nonumber\\
    &- p(f_t(\bx)\leq y) \mathbb{E}_{p(\bx|f_t(\bx)\leq y)}\Big[\bg(f_t(\bx))\Big].
\end{align}

Finally, from Lemma~\ref{lemma_obsprocess_E}, we can rewrite Eq.~\eqref{replacedGradient_new} as
\begin{align}
    \nabla \mathcal{L}(f_t) =&p(f_t(\bx)\leq y) \mathbb{E}_{p(\bx,y'|f_t(\bx)\leq y')}[\nabla L(f_t(\bx),y)]\nonumber\\
    &+ \mathbb{E}_{p(\bx)}\Big[\bg(f_t(\bx))\Big]\nonumber\\
    &- p(f_t(\bx)\leq y) \mathbb{E}_{p(\bx|f_t(\bx)\leq y')}\Big[\bg(f_t(\bx))\Big],
\end{align}
which is identical to Eq.~\eqref{finalGradient_new}.
Thus, the gradient in Eq.~\eqref{finalGradient_new} is unbiased and consistent with the gradient in Eq.~\eqref{gradient_new} a.s.
\end{proof}

\subsection{Proof of Lemma~\ref{naive}}
\label{ap_pr_naive}
\begin{proof}
The difference between the decomposed gradients $\nabla \check{\mathcal{L}}(f_t)$ and $\nabla \mathcal{L}(f_t)$ at step $t+1$ in the gradient descent is
\begin{align}
    |\nabla \check{\mathcal{L}}(f_t)&-\nabla \mathcal{L}(f_t)|\\\nonumber
    =&\bigg|p(f_t(\bx)\leq y) \mathbb{E}_{p(\bx,y'|f_t(\bx)\leq y')}[\nabla L(f_t(\bx),y)]\\\nonumber
    &+ p(y<f_t(\bx)) \mathbb{E}_{p(\bx,y'|y'<f_t(\bx))}[\nabla L(f_t(\bx),y)]\\\nonumber
    -&p(f_t(\bx)\leq y) \mathbb{E}_{p(\bx,y|f_t(\bx)\leq y)}[\nabla L(f_t(\bx),y)]\\\nonumber
    &- p(y<f_t(\bx)) \mathbb{E}_{p(\bx,y|y<f_t(\bx))}[\nabla L(f_t(\bx),y)]\bigg|.
\end{align}
From Lemma~\ref{lemma_obsprocess_E} and Condition~\ref{condition_L},
\begin{align}
    |\nabla \check{\mathcal{L}}(f_t)-&\nabla \mathcal{L}(f_t)|\\\nonumber
    =&\bigg|p(y<f_t(\bx)) \mathbb{E}_{p(\bx,y'|y'<f_t(\bx))}[\nabla L(f_t(\bx),y)]\\\nonumber
    &- p(y<f_t(\bx)) \mathbb{E}_{p(\bx,y|y<f_t(\bx))}[\nabla L(f_t(\bx),y)]\bigg|\\\nonumber
    =&\bigg|p(y<f_t(\bx)) \mathbb{E}_{p(\bx|y'<f_t(\bx))}[\bg(f_t(\bx))]\\\nonumber
    &- p(y<f_t(\bx)) \mathbb{E}_{p(\bx|y<f_t(\bx))}[\bg(f_t(\bx))]\bigg|.
\end{align}
We decompose $\mathbb{E}_{p(\bx|y'<f_t(\bx))}[\bg(f_t(\bx))]$ again as
\begin{align}
    &|\nabla \check{\mathcal{L}}(f_t)-\nabla \mathcal{L}(f_t)|\\\nonumber
    &=\bigg|p(y<f_t(\bx)) \bigg(\\\nonumber
    &p(f_t(\bx)\leq y|y'<f_t(\bx)) \mathbb{E}_{p(\bx|y'<f_t(\bx) \land f_t(\bx)\leq y)}[\bg(f_t(\bx))]\\\nonumber
    &+p(y<f_t(\bx)|y'<f_t(\bx))\\\nonumber
    &~~~~~~~~~~~~~~~~~~~~~~~~~~~~~~~~~~~~\mathbb{E}_{p(\bx|y'<f_t(\bx) \land y<f_t(\bx))}[\bg(f_t(\bx))]\bigg)\\\nonumber
    &~~~~~~~~~~~~~~~~~~~~~~~~~~- p(y<f_t(\bx)) \mathbb{E}_{p(\bx|y<f_t(\bx))}[\bg(f_t(\bx))]\bigg|.
\end{align}
The condition $y'<f_t(x)\land y<f_t(x)$ is equivalent to the condition $y<f_t(x)$ since $y'\leq y$ from Assumption~\ref{assumption_obsprocess}. Then, we have
\begin{align}
    &|\nabla \check{\mathcal{L}}(f_t)-\nabla \mathcal{L}(f_t)|\\\nonumber
    &=\bigg|p(y<f_t(\bx)) \bigg(\\\nonumber
    &p(f_t(\bx)\leq y|y'<f_t(\bx))\mathbb{E}_{p(\bx|y'<f_t(\bx) \land f_t(\bx)\leq y)}[\bg(f_t(\bx))]\\\nonumber
    &+p(y<f_t(\bx)|y'<f_t(\bx))\\\nonumber
    &~~~~~~~~~~~~~~~~~~~~~~~~~~~~~~~~~~~~\mathbb{E}_{p(\bx|y<f_t(x))}[\bg(f_t(\bx))]\bigg)\\\nonumber
    &~~~~~~~~~~~~~~~~~~~~~~~~~~- p(y<f_t(\bx)) \mathbb{E}_{p(\bx|y<f_t(\bx))}[\bg(f_t(\bx))]\bigg|.
\end{align}
Additionally, since $p(y<f_t(\bx)|y'<f_t(\bx))=1-p(f_t(\bx)\leq y|y'<f_t(\bx))$,
\begin{align}
    &|\nabla \check{\mathcal{L}}(f_t)-\nabla \mathcal{L}(f_t)|\\\nonumber
    &=\bigg|p(y<f_t(\bx))p(f_t(\bx)\leq y|y'<f_t(\bx))\bigg(\\\nonumber
    &~~~~~~~~~~~~~~~~~~~~~~~~~~~~~~~~~~~~~~~~\mathbb{E}_{p(\bx|y'<f_t(\bx) \land f_t(\bx)\leq y)}[\bg(f_t(\bx))]\\\nonumber
    &~~~~~~~~~~~~~~~~~~~~~~~~~~~~~~~~~~~~~~~~~~~~~-\mathbb{E}_{p(\bx|y<f_t(\bx))}[\bg(f_t(\bx))]\bigg)\bigg|.
\end{align}
\noindent This equation shows that the bias is represented by the difference between the expectation of $\bg(f_t(\bx))$ with the lower-side data and that with the original upper-side data mixed into the lower side due to the asymmetric noise and the corresponding proportions.

From Assumption~\ref{additional_es}, since $\epsilon_{\mathrm{a}} \indep \bx$,
\begin{align}
    &|\nabla \check{\mathcal{L}}(f_t)-\nabla \mathcal{L}(f_t)|\\\nonumber
    &=\bigg|p(y<f_t(\bx))p(f_t(\bx)\leq y|y'<f_t(\bx))\bigg(\\\nonumber
    &\mathbb{E}_{p(\bx|f_t(\bx)\leq y)}[\bg(f_t(\bx))]-\mathbb{E}_{p(\bx|y<f_t(\bx))}[\bg(f_t(\bx))]\bigg)\bigg|.
\end{align}
Since $|f-f^*| \leq |\epsilon_{\mathrm{s}}|$ a.s., $p(f_t(\bx)\leq y)=\eta$ and $p(y<f_t(\bx))=1-\eta$ from their definition,
\begin{align}
    &p(f_t(\bx)\leq y|y'<f_t(\bx))=\nonumber\\
    &~~~~~~~~~~~~~~~~~~~~~~~~~~~~\frac{p(f_t(\bx)\leq y)p(\epsilon_{\mathrm{a}}<0)}{p(y<f_t(\bx))+p(f_t(\bx)\leq y)p(\epsilon_{\mathrm{a}}<0)}\nonumber\\
    &~~~~~~~~~~~~~~~~~~~~~~~~~~~~~~~~~~~~~~~~=\frac{\eta(1-\xi)}{(1-\eta)+\eta(1-\xi)}.
\end{align}
Therefore, from the definition of $\delta$,
\begin{align}
    |\nabla \check{\mathcal{L}}(f_t)-\nabla \mathcal{L}(f_t)| &\geq \frac{\eta(1-\eta)(1-\xi)}{(1-\eta)+\eta(1-\xi)}\delta\nonumber\\
    &=\frac{\eta(1-\eta)(1-\xi)}{1-\eta\xi}\delta.
\end{align}
\end{proof}

\section{Implementation of Learning Algorithm Based on Stochastic Optimization}
\label{ap_implementation}
We scale up our U2 regression algorithm by stochastic approximation with $M$ mini-batches and add a regularization term, $R(f)$:
\begin{align}
    \label{mini-batchgradient}
    &\nabla \mathcal{\hL}^{\{m\}}(f_t)=\sum_{(\bx,y) \in \big\{\bX_{\mathrm{up}}^{\{m\}}, \by_{\mathrm{up}}^{\{m\}}\big\}} \nabla L(f_t(\bx),y)\\\nonumber
    &+ \rho \Bigg[\sum_{\bx \in \bX_{\mathrm{un}}^{\{m\}}} \bg(f_t(\bx)) \Bigg]- \!\sum_{\bx \in \bX_{\mathrm{up}}^{\{m\}}} \bg(f_t(\bx)) + \lambda \frac{\partial R(f_t)}{\partial \btheta},
\end{align}
where $\nabla \mathcal{\hL}^{\{m\}}(f_t)$ is the gradient for the $m$-th mini-batch, $\{\bX_{\mathrm{up}}^{\{m\}}, \by_{\mathrm{up}}^{\{m\}}\}$ and $\bX_{\mathrm{un}}^{\{m\}}$ are the upper-side and unlabeled sets in the $m$-th mini-batch based on the current $f_t$, respectively, $\lambda$ is a regularization parameter, and the regularization term $R(f)$ is, for example, the L$1$ or L$2$ norm of the parameter vector $\btheta$ of $f$. 
We also convert $\nicefrac{n_{\mathrm{up}}}{(\pi_{\mathrm{up}} N)}$ to a hyperparameter $\rho$, ignoring constant coefficients instead of directly handling $\pi_{\mathrm{up}}$. The hyperparameters $\rho$ and $\lambda$ are optimized in training based on the grid-search with the validation set.

The U2 regression algorithm based on stochastic optimization is described in Algorithm~\ref{alg1}.
We learn the regression function with the gradient in Eq.~\eqref{mini-batchgradient} by utilizing any stochastic gradient method.
By using the learned $f$, we can estimate $\hy = f(\bx)$ for new data $\bx$.

\section{Algorithm for LU Regression}
\label{ap_LU}
We show the algorithm for the \emph{lower and unlabeled} regression (LU regression), where labels for some observations may become inconsistently \emph{higher} than those for typical observations.
Let $L_{\mathrm{LU}}(f(\bx),y)$ be a loss function for LU regression and $\bg_{\mathrm{LU}}(f(\bx))$ be a gradient of $\nabla L_{\mathrm{LU}}(f(\bx),y)$ when $f(\bx)\leq y$.
Similar to Condition~\ref{condition_L} for U2 regression, we assume that the class of $L_{\mathrm{LU}}(f(\bx),y)$ satisfies the condition that $\bg_{\mathrm{LU}}(f(\bx))$ is a gradient function depending only on $f(\bx)$ and not on the value of $y$.
Then, LU regression is derived as Algorithm~1, with the following gradient, $\nabla \mathcal{\hat{L}}_{\mathrm{LU}}^{\{m\}}(f_t)$, instead of $\nabla \mathcal{\hat{L}}^{\{m\}}(f_t)$ in Eq.~\eqref{mini-batchgradient}, as
\begin{align}
    \label{mini-batchgradient_LU}
    \nabla \mathcal{\hat{L}}_{\mathrm{LU}}^{\{m\}}(f_t)=& \sum_{\{\bx,y\} \in \big\{\bX_{\mathrm{lo}}^{\{m\}}, \by_{\mathrm{lo}}^{\{m\}}\big\}} \nabla L_{\mathrm{LU}}(f_t(\bx),y)\\\nonumber
    &+ \rho \Bigg[\sum_{\bx \in \bX_{\mathrm{un}}^{\{m\}}} \bg_{\mathrm{LU}}(f_t(\bx)) \Bigg]\\\nonumber
    &- \sum_{\bx \in \bX_{\mathrm{lo}}^{\{m\}}} \bg_{\mathrm{LU}}(f_t(\bx))+ \lambda \frac{\partial R(f_t)}{\partial \btheta},
\end{align}
where $\{\bX_{\mathrm{lo}}^{\{m\}}, \by_{\mathrm{lo}}^{\{m\}}\}$ and $\bX_{\mathrm{un}}^{\{m\}}$ are the lower-side and unlabeled sets in the $m$-th mini-batch based on the current $f_t$, respectively.
\begin{algorithm}[t]
\caption{U2 regression based on stochastic gradient method.}
\label{alg1}
\begin{algorithmic}[1]
    \REQUIRE Training data $\calD'= \{\bx_{n},y'_{n} \}^N_{n=1}$;
    hyperparameters $\rho, \lambda \geq 0$; an external stochastic gradient method $\mathcal{A}$
    \ENSURE Model parameters $\btheta$ for $f$
    \WHILE{No stopping criterion has been met}
    \STATE Shuffle $\calD'$ into $M$ mini-batches: $\big\{\bX^{\{m\}}, \by^{\{m\}}\big\}_{m=1}^M$
    \FOR{$m=1$ \textbf{to} $M$}
    \STATE Compute the gradient $\nabla \hat{\mathcal{L}}^{\{m\}}(f_t)$ in Eq.~\eqref{mini-batchgradient} with $\big\{\bX^{\{m\}}, \by^{\{m\}}\big\}$
    \STATE Update $\btheta$ by $\mathcal{A}$ with $\nabla \hat{\mathcal{L}}^{\{m\}}(f_t)$
    \ENDFOR
    \ENDWHILE
\end{algorithmic}
\end{algorithm}

\section{Computing Infrastructure}
\label{ap_comp}
All of the experiments were carried out with a Python and TensorFlow implementation on workstations having $80$~GB of memory, a $4.0$~GHz CPU, and an Nvidia Titan X GPU. In this environment, the computational time to produce the results was a few hours.
\begin{figure*}[t]
	\centering
	\includegraphics[width=170mm]{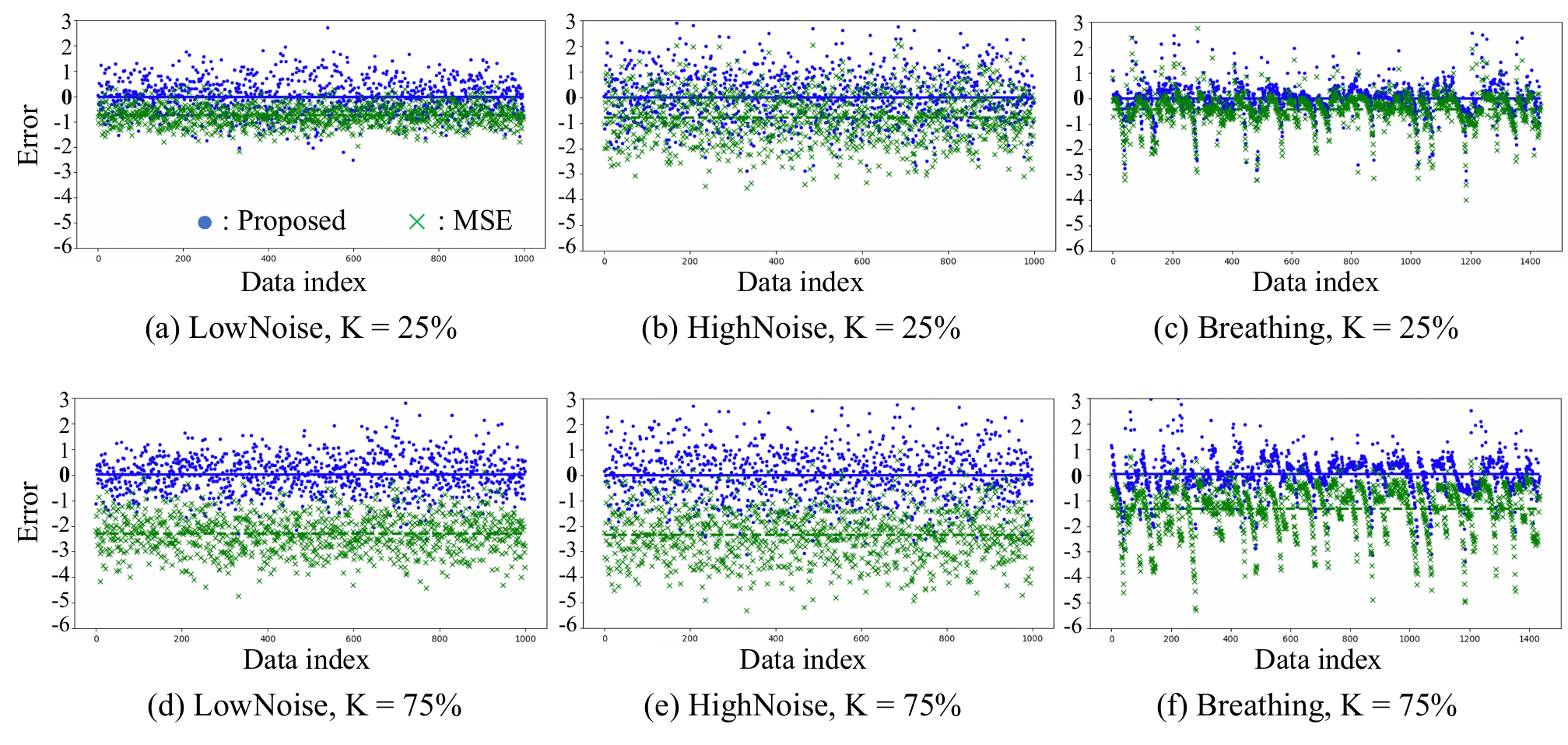}
	\caption{Errors in prediction (predicted value minus true value) by proposed method (blue) and by \texttt{MSE} (green) for tasks (a) \textbf{LowNoise}, $K = 25 \%$, (b) \textbf{HighNoise}, $K = 25 \%$, (c) \textbf{Breathing}, $K = 75 \%$, (d) \textbf{LowNoise}, $K = 75 \%$, (e) \textbf{HighNoise}, $K = 75 \%$, and (f) \textbf{Breathing}, $K = 75 \%$. Error of each data point is shown by a dot (for proposed method) or a cross mark (for \texttt{MSE}), and average error is shown by a solid line (for proposed method) or dashed line (for \texttt{MSE}).}
	\label{FigResults2575}
\end{figure*}

\section{Details of Experiments in Section~\ref{sec:demo}}
\label{ap_ex1}
\subsection{Details of \textbf{LowNoise} and \textbf{HighNoise} tasks}
We conducted the experiments on synthetic data to evaluate the feasibility of our method for obtaining unbiased learning results from asymmetrically corrupted data containing different proportions of incomplete observations. We generated synthetic data on the basis of Assumption~\ref{assumption_obsprocess} and Eq.~\eqref{observation_process}.
We randomly generated $N=1,000$ training samples, $\bX = \{\bx_n\}_{n=1}^{N}$, from the standard Gaussian distribution $\NormalDist(\bx_n ; 0, \bI)$, where the number of features in $\bx$ was $D=10$, and $\bI$ is the identity matrix. Then, using $\bX$, we generated the corresponding $N$ sets of true labels $\by = \{y_n\}_{n=1}^{N}$ from the distribution $\NormalDist(y_n ;\bw^\top \bx_n, \beta)$, where $\bw$ are coefficients that were also randomly generated from the standard Gaussian distribution $\NormalDist(\bw ; 0, \bI)$, $\beta$ is the noise precision, and $\top$ denotes the transpose. For simulating the situation in which a label has incomplete observations, we created corrupted labels $\by'= \{y'_n\}_{n=1}^{N}$ by randomly selecting the $K$ percent of data in $\by$ and subtracting the absolute value of white Gaussian noise with the standard deviation having twice the value of that for generating $\by$ from their values.
We repeatedly evaluated the proposed method for each of the following settings. The noise precision was $\beta = \{10^0,10^{-1}\}$, which corresponds to a low-noise setting task (\textbf{LowNoise}) and a high-noise setting task (\textbf{HighNoise}), and the proportion of incomplete training samples was $K = \{ 25, 50, 75 \}\%$. In the case of $K=75\%$, only $25$ percent of the samples correctly corresponded to labels, and all of the other samples were attached with labels that were lower than the corresponding true values. It is quite difficult to learn regression functions using such data.

\paragraph{Implementation.}
In these tasks, we used a linear model, $\btheta^\top \bx$, for $f(\bx)$ and an implementation for Eq.~\eqref{mini-batchgradient} with the absolute loss, which satisfies Condition~\ref{condition_L}, for the loss function $L$ and L$1$-regularization for the regularization term.
We set the candidates of the hyperparameters, $\rho$ and $\lambda$, to $\{10^{-3},10^{-2},10^{-1},10^{0} \}$. We standardized the data by subtracting their mean and dividing by their standard deviation in the training split. We used Adam with the hyperparameters recommended in~\cite{kingma2014adam}, and the number of samples in the mini-batches was set to $32$.

\subsection{Details of \textbf{Breathing} Task}
We also used a real-world sensor dataset collected from the Kaggle dataset~\citep{SagarSen2016} that contains breathing signals (\textbf{Breathing}). The dataset consisted of $N=1,432$ samples. We utilized signals from a chest belt as $\bX = \{\bx_n\}_{n=1}^{N}$, and $\bx$ in each sample had $D=2$ number of features, i.e., the period and height of the expansion/contraction of the chest. We utilized signals obtained by the Douglas bag (DB) method, which is the gold standard for measuring ventilation, as true labels $\by = \{y_n\}_{n=1}^{N}$. For our problem setting, we created corrupted labels $\by'= \{y'_n\}_{n=1}^{N}$ through the same procedure for synthetic corruption as that for LowNoise and HighNoise with $K = \{ 25, 50, 75 \}\%$.

\paragraph{Implementation.}
In the experiment on Breathing, for its non-linearity, we used $\btheta^\top \bphi(\bx,\sigma)$ for $f(\bx)$, where $\bphi$ is a radial basis function with the training set as its bases, and $\sigma$ is a hyperparameter representing the kernel width that is also optimized by using the validation set. We set the candidates of the hyperparameter $\sigma$ to $\{10^{-3},10^{-2},10^{-1},10^{0}\}$. The other implementation details were the same as those for \textbf{LowNoise} and \textbf{HighNoise}.

\subsection{Detailed Results}
Figure~\ref{FigResults2575} shows the error between the estimation results of the proposed method and their true values and those of \texttt{MSE} for \textbf{LowNoise}, \textbf{HighNoise}, and \textbf{Breathing} with $25$ and $75$ percent of incomplete training samples.
Table~\ref{ResultSD} shows the performance on \textbf{LowNoise}, \textbf{HighNoise}, and \textbf{Breathing} for the proposed method and \texttt{MSE}.
As shown in Fig.~\ref{FigResults2575}, the proposed method obtained unbiased learning results in all cases, while \texttt{MSE} produced biased results.
From Table~\ref{ResultSD}, we can see that the proposed method outperformed \texttt{MSE} overall with $0<K$. We found that the performance of our method was not significantly affected by the increase in the proportion of incomplete training samples $K$ even for $K=75\%$, unlike that of \texttt{MSE}. We also show results with $K=0 \%$, which means there is no corruption. Except for the \textbf{LowNoise} task, the proposed method worked comparably to MSE even when $K=0 \%$, which shows that there is almost no drawback if we use the proposed method even for the uncorrupted data without incomplete observations. Since \textbf{LowNoise} with $K=0 \%$ is very clean and simple data with a linear relationship between $\bx$ and $y$ and little additive white Gaussian noise, MSE can perform much better than robust methods, including the proposed method. We again note that this paper addresses cases where there is the asymmetric label corruption. If there is no corruption, we can use simple loss functions, such as squared loss.
\begin{table*}[t]
\caption{Comparison of proposed method and \texttt{MSE} in terms of MAE (smaller is better). Best methods are in bold. Confidence intervals are standard errors.}
\label{ResultSD}
\centering
\begin{tabular}{ccccc}
    \\[-6pt]
    \multicolumn{5}{c}{(a) LowNoise}\\
    \cmidrule(lr){1-5}
    & $K=0\%$&$K=25\%$&$K=50\%$&$K=75\%$\\
    \midrule
     \texttt{MSE}& $\bm{0.24  \pm  0.01}$& $0.77  \pm  0.01$& $1.53  \pm  0.02$& $2.30  \pm  0.02$\\
     Proposed& $0.57  \pm  0.01$& $\bm{0.55  \pm  0.01}$& $\bm{0.54  \pm  0.01}$& $\bm{0.58  \pm  0.01}$\\
    \bottomrule
\end{tabular}
\begin{tabular}{ccccc}
    \\[-1pt]
    \multicolumn{5}{c}{(b) HighNoise}\\
    \cmidrule(lr){1-5}
    & $K=0\%$&$K=25\%$&$K=50\%$&$K=75\%$\\
    \midrule
     \texttt{MSE}& $0.77  \pm  0.02$ &$1.03  \pm  0.02$& $1.62  \pm  0.03$& $2.36  \pm  0.03 $\\
     Proposed& $0.79  \pm  0.02$ & $\bm{0.79  \pm  0.02}$& $\bm{0.80  \pm  0.02}$& $\bm{0.80  \pm  0.02} $\\
    \bottomrule
\end{tabular}
\begin{tabular}{ccccc}
    \\[-1pt]
    \multicolumn{5}{c}{(c) Breathing}\\
    \cmidrule(lr){1-5}
    & $K=0\%$&$K=25\%$&$K=50\%$&$K=75\%$\\
    \midrule
    \texttt{MSE}& $0.41  \pm  0.02$& $0.55  \pm  0.02$& $0.91  \pm  0.02$& $1.32  \pm  0.02$\\
    Proposed& $0.45  \pm  0.01$& $\bm{0.43  \pm  0.01}$& $\bm{0.46  \pm  0.01}$& $\bm{0.59  \pm  0.01}$\\
    \bottomrule
\end{tabular}
\end{table*}

\section{Details of Experiments in Section~\ref{sec:diffsize}}
\label{ap_exdiffsize}
\subsection{Detailed Results}
\begin{figure*}[t]
	\centering
	\includegraphics[width=170mm]{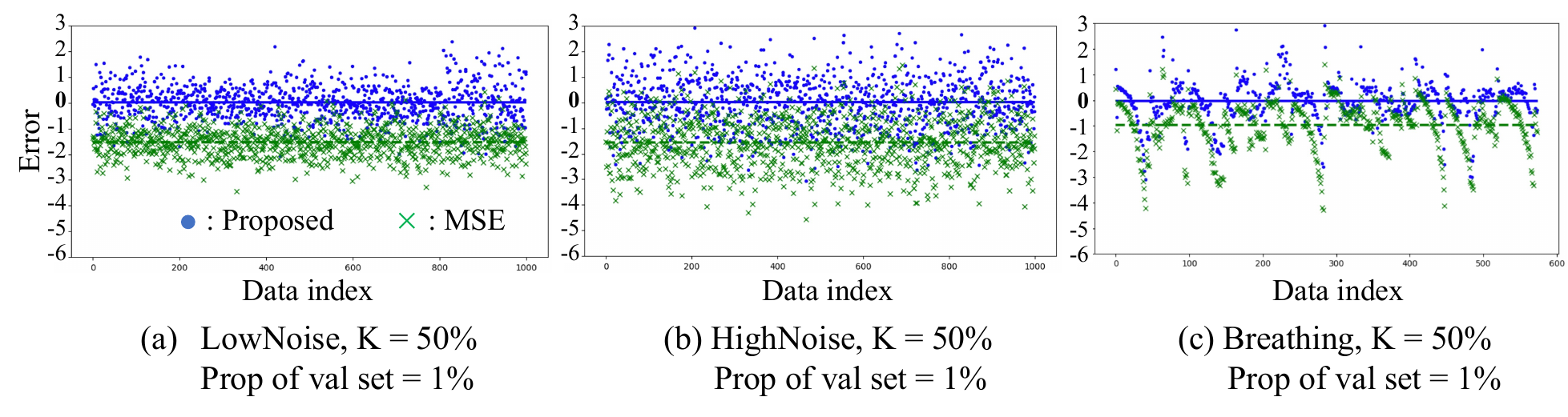}
	\caption{Errors in prediction when we choose hyperparameters with $1 \%$ of the training set as a validation set. Other configurations were the same as those in Fig.~\ref{FigResultsSD}.}
	\label{FigResults1pval}
\end{figure*}

In Fig.~\ref{FigResults1pval}, we show charts similar to Fig.~\ref{FigResultsSD} (the error in prediction for \textbf{LowNoise}, \textbf{HighNoise}, and \textbf{Breathing} tasks with $K = 50 \%$) when we used $1 \%$ of the training set as the validation set. We can see that even in this case, the proposed method achieved unbiased learning (the average error shown by the blue solid line is approximately zero).

\section{Details of Experiments in Section~\ref{sec:performance}}
\label{ap_ex2}
\subsection{Details of Task}
We applied the algorithm to five different real-world healthcare tasks recorded in the datasets from the UCI Machine Learning Repository~\citep{Velloso2013,velloso2013qualitative},
which contains sensor outputs from wearable devices attached to the arm while subjects exercised.
From the non-intrusive sensors attached to gym equipment, we estimated the motion intensity of a subject that was measured accurately with an intrusive sensor wrapped around the arm. If we can mimic outputs from the arm sensor with outputs from the equipment sensor, it could contribute to the subjects' comfort, as they would not need to wear sensors to measure their motion intensity. 
We utilized all of the features from the equipment sensor that took ``None" values less than ten times as $\bX = \{\bx_n\}_{n=1}^{N}$, where each sample had $D=13$ number of features. The corrupted labels $\by' = \{y'_n\}_{n=1}^{N}$ were the magnitude of acceleration from the arm sensor, which can accurately sense motion intensity on the arm, but it had insufficient data coverage and incomplete observations for the movements of other body parts.
For performance evaluation, we used the magnitude of acceleration for the entire body as true labels $\by = \{y_n\}_{n=1}^{N}$. The number of samples was $N=11,159$, $N=7,593$, $N=6,844$, $N=6,432$, and $N=7,214$ for the \textbf{Specification}, \textbf{Throwing A}, \textbf{Lifting}, \textbf{Lowering}, and \textbf{Throwing B} tasks, respectively.

\paragraph{Implementation.}
For the complex nature of the tasks, we used a $6$-layer multilayer perceptron with ReLU~\citep{nair2010rectified} (more specifically, $D$-$100$-$100$-$100$-$100$-$1$) as $f(\bx)$, which demonstrates the usefulness of the proposed method for training deep neural networks. We also used a dropout~\citep{srivastava2014dropout} with a rate of $50\%$ after each fully connected layer.
We used two implementations of $L(f(\bx),y)$ for $f(\bx)\leq y'$ in Eq.~\eqref{empiricalgradient}: the absolute loss ({\tt Proposed-1}) and the squared loss ({\tt Proposed-2}). In both implementations, we use the absolute loss, which satisfies Condition~\ref{condition_L}, for $L(f(\bx),y)$ when $y'<f(\bx)$. We used L$1$-regularization for the regularization term. The other implementation details were the same as those for \textbf{LowNoise}, \textbf{HighNoise}, and \textbf{Breathing}.

\section{Details of Experiments in Section~\ref{sec:usecase}}
\label{ap_ex3}
\subsection{Details of Task}
We demonstrate the practicality of our approach in a real use case in healthcare. From non-intrusive bed sensors installed under each of the four legs of a bed, we estimated the motion intensity of a subject that was measured accurately with ActiGraph, a gold standard intrusive sensor wrapped around the wrist~\citep{tryon2013activity,mullaney1980wrist,webster1982activity,cole1992automatic}. The sensing results of ActiGraph are used for tasks such as discriminating whether a subject is asleep or awake~\citep{cole1992automatic}.
While ActiGraph can accurately sense motion on the forearm, it has insufficient data coverage in other areas and often causes observations of movements on other body parts to be missing.
The bed sensors have a broader data coverage since they can sense global motion on all body parts; however, the sensing accuracy is limited due to their non-intrusiveness.
If we can mimic the outputs from ActiGraph with outputs from the bed sensors, we can expect to achieve sufficient accuracy and coverage while also easing the burden on the subject.
The dataset we used included three pieces of data, Data (i), (ii), and (iii), which were respectively recorded over $20$, $18$, and $18.5$ minutes. Each piece of data consisted of pairs of bed-sensor-data sequences and the corresponding motion intensity sequence obtained by ActiGraph. We used the ``magnitude'' attribute of ActiGraph as corrupted labels $\by'$ for the motion intensity, whose sampling rate was about one sample per second. For true labels $\by$, we manually measured the motion intensity every minute under the management of a domain expert. For $\bX$, we first computed the gravity center of the four sensor outputs that were obtained from the bed sensors under the four legs of a bed. Then, we computed the time derivatives and cross terms of the raw sensor outputs and the gravity center. The sampling rate of the bed sensors was different from that of ActiGraph (about one sample per five milliseconds). Thus, $\bX$ was finally generated as a sliding window of statistics in $1,000$-millisecond ($1$-second) subsequences of the time series of the above computed variables, where $1$ second was the same as the sampling interval of ActiGraph. The statistics were means, standard deviations, and $\{0.05, 0.25, 0.5, 0.75, 0.95\}$ quantiles. In the end, the numbers of samples and features were $N = 3,390$ and $D = 84$.

\paragraph{Implementation.}
In this task, we used the linear model $\btheta^\top \bx$ for $f(\bx)$ due to its interpretability, which is inevitable in real-world healthcare and medical applications. The other implementation details were the same as those for \textbf{LowNoise} and \textbf{HighNoise}.

\subsection{Estimation Results for Motion Intensity}
\begin{figure}[t]
	\centering
	\includegraphics[width=80mm]{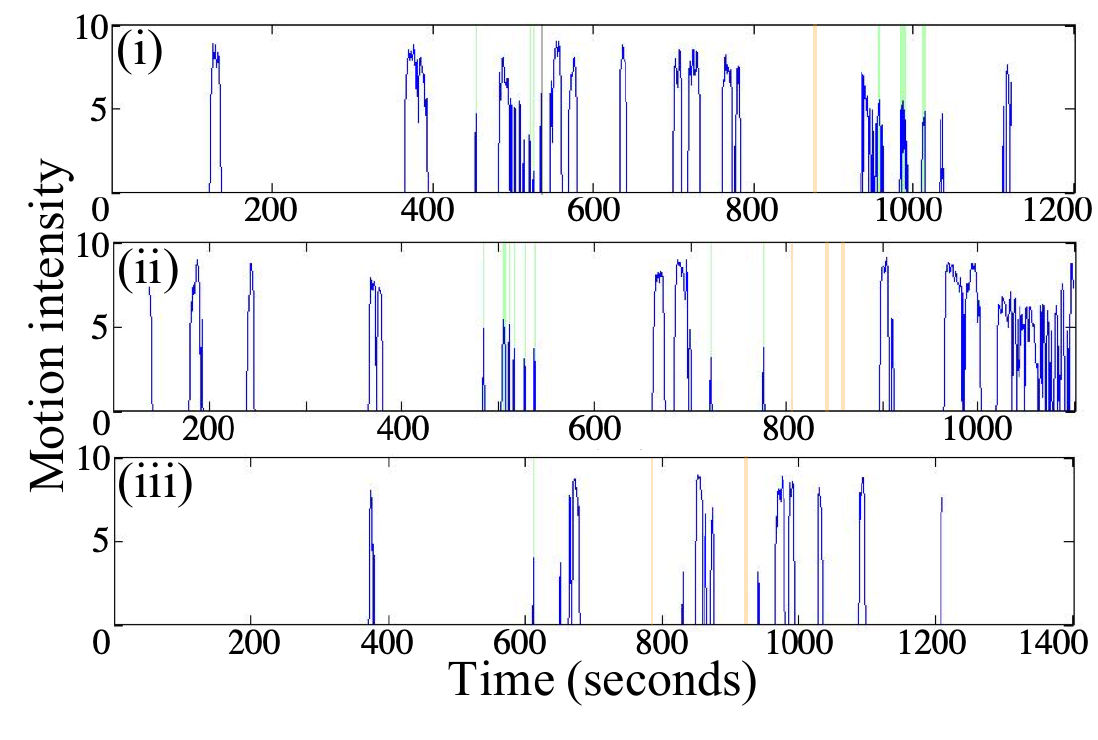}
	\caption{Experiment on real use case for healthcare. Blue line represents our estimation results for motion intensity. White (non-colored) area shows that both proposed method and ActiGraph correctly estimated motion intensity of the subject at this duration. Green area shows that our method could capture motion at this duration while ActiGraph could not. Orange area shows that our method could not capture motion at this duration but ActiGraph could. Gray area shows that our method mistakenly captured noise as subject's motion.}
	\label{FigResultCase}
\end{figure}

Figure~\ref{FigResultCase} compares our estimation results for motion intensity with the output of ActiGraph and true labels.

\subsection{Important Features for Estimating Motion Intensity}
The important features selected by L$1$ regularization were the statistics of the gravity center and the cross terms and time derivatives of the raw sensor outputs. The largest weight was assigned to the standard deviation of the gravity center, which represents the amplitude of the gravity center, so it is directly related to the motion of subjects.

\end{document}